\documentclass[11pt]{article}

\usepackage{style/iclr2026_conference}
\usepackage{xcolor}
\usepackage{times}
\usepackage{microtype}
\usepackage{graphicx}
\usepackage{subcaption}
\usepackage{float}
\usepackage{booktabs}
\usepackage{multirow}
\usepackage{makecell}
\usepackage{threeparttable}
\usepackage{tabularx}
\usepackage{tcolorbox}
\tcbuselibrary{breakable,skins}
\usepackage{fvextra}
\usepackage{siunitx}
\usepackage{longtable}
\usepackage{array}
\usepackage{amsmath}
\usepackage{amssymb}
\usepackage{mathtools}
\usepackage{xspace}
\usepackage{xcolor}
\usepackage{colortbl}
\usepackage{enumitem}
\usepackage{url}
\usepackage{xurl}
\usepackage[colorlinks=true,allcolors=blue!45!black]{hyperref}
\usepackage[capitalise,noabbrev]{cleveref}

\definecolor{copawblue}{HTML}{315C9B}
\definecolor{copawgreen}{HTML}{207A4A}
\definecolor{copawgray}{HTML}{666666}
\definecolor{copawtodo}{HTML}{B42318}
\definecolor{copawgroupbg}{HTML}{F2F2F2}
\definecolor{counttagbg}{HTML}{E8F0F5}
\definecolor{counttagfg}{HTML}{355A6B}
\definecolor{ratetagbg}{HTML}{E8F2ED}
\definecolor{ratetagfg}{HTML}{426B5D}
\definecolor{stattagbg}{HTML}{F6EEE4}
\definecolor{stattagfg}{HTML}{805E3F}
\definecolor{abstractpurple}{HTML}{6243A5}
\definecolor{copawpromptbg}{HTML}{F5F7FA}
\definecolor{copawprompttitlebg}{HTML}{E7EDF5}
\definecolor{copawpromptframe}{HTML}{8EA3BD}

\renewenvironment{abstract}{%
  \begin{tcolorbox}[
    colback=white,
    colframe=abstractpurple,
    boxrule=0.8pt,
    arc=2.5mm,
    outer arc=2.5mm,
    left=5mm,
    right=5mm,
    top=2.5mm,
    bottom=3.5mm,
    before skip=0.075in,
    after skip=1ex
  ]%
  \begin{center}
    \large\bfseries\color{abstractpurple} Abstract
  \end{center}
  \vspace{-0.6ex}
  \normalfont\color{black}\noindent
}{%
  \end{tcolorbox}%
}

\crefname{section}{Section}{Sections}
\crefname{figure}{Figure}{Figures}
\crefname{table}{Table}{Tables}
\crefname{appendix}{Appendix}{Appendices}

\newcommand{\frameworkname}{Cowork-Gym\xspace}

\newcommand{\papertitle}{%
  \frameworkname: Synthesizing Interactive Work Environments at Scale%
}

\title{\centering\papertitle}

\makeatletter
\patchcmd{\@maketitle}
  {{\LARGE\sc \@title\par}}
  {\vspace*{-0.8cm}{\LARGE\sc \@title\par}\vspace{0.3cm}}
  {}{\PackageError{paper_info}{Could not adjust the title spacing}{}}
\makeatother

\newcommand{\authornotetext}{%
  \begingroup
  \renewcommand{\thefootnote}{\textdagger}%
  \footnotetext[1]{Equal contribution.}%
  \renewcommand{\thefootnote}{*}%
  \footnotetext[1]{Corresponding authors.}%
  \endgroup
}

\author{%
  \begin{minipage}{0.96\textwidth}
    \centering
    Anni Zou\textsuperscript{\textdagger}, Cheng Chen\textsuperscript{\textdagger}, Jiaji Deng\textsuperscript{\textdagger}, Weikang Zhou\textsuperscript{\textdagger}, Xinji Mai\textsuperscript{\textdagger}, Yue Wu\textsuperscript{\textdagger},\\
    Jianing Lin, Lipeng Xie, Lixiong Qin, Qingyue Zhao, Que Shen, Shuchang Tao, \\ Xuchen Pan, Yaliang Li, Yunpeng Zhai, Yushuo Chen, Zhihai Wang, Zhuo Zhang, \\ Zhaoyang Liu\textsuperscript{*}, Bolin Ding\textsuperscript{*}, Jingren Zhou\\[0.55em]
    \textnormal{\raisebox{-0.22em}{\includegraphics[height=1.2em]{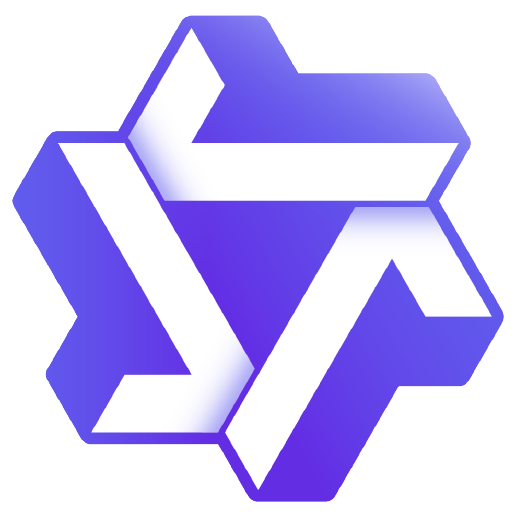}}\hspace{0.35em}%
    \textit{Alibaba Token Foundry}}\\[0.4em]
    {\small\normalfont\fontfamily{pcr}\selectfont
    \{jingmu.lzy, bolin.ding\}@alibaba-inc.com}
  \end{minipage}
}

\newcommand{\gainbadge}[1]{%
  \begingroup
  \def\gainshade{10}%
  \ifdim#1pt>8pt
    \def\gainshade{32}%
  \else
    \ifdim#1pt>5pt
      \def\gainshade{21}%
    \fi
  \fi
  \setlength{\fboxsep}{1.6pt}%
  \colorbox{copawgreen!\gainshade}{\textcolor{copawgreen!85!black}{\scriptsize\bfseries +#1}}%
  \endgroup
}

\newcommand{\seentag}{%
  \begingroup
  \setlength{\fboxsep}{1.4pt}%
  \colorbox{copawblue!12}{\textcolor{copawblue}{\scriptsize\bfseries SEEN}}%
  \endgroup
}
\newcommand{\unseentag}{%
  \begingroup
  \setlength{\fboxsep}{1.4pt}%
  \fcolorbox{copawgray!60}{white}{\textcolor{copawgray}{\scriptsize\bfseries UNSEEN}}%
  \endgroup
}

\newif\ifusevone%
\usevonefalse%

\iclrfinalcopy{}

\begin{document}

\maketitle
\authornotetext
\lhead{\includegraphics[height=0.9em]{figures/logo.pdf}}

\begin{abstract}
Cowork agents navigate dependencies across tools and files while adapting to changing user requirements and edits to shared workspaces. Scaling such environments requires maintaining consistency among task objectives, executable resources, user behavior, and success criteria.
We introduce \frameworkname, a framework that synthesizes Cowork
environments through a build--execute--revise loop that uses execution
evidence and review feedback to validate and repair generated environments.
To capture realistic Cowork dynamics, a stateful user simulator responds to
the agent's actions through dialogue and direct edits to the shared workspace.
We synthesize 64.9K environments across 14 application scenarios and use them to post-train 9B, 27B, and 35B-A3B Qwen models with supervised fine-tuning, reinforcement learning, and on-policy distillation. Averaged across six agentic benchmarks, these models outperform their corresponding base checkpoints by 8.7, 6.4, and 4.1 points, respectively. Ours-27B ranks second overall among the evaluated models and achieves the highest average scores on PawBench and ClawEval, outperforming models with over 100B total parameters on both benchmarks.

\begin{flushleft}
\urlstyle{same}
\raisebox{-0.22em}{%
  \includegraphics[height=1.15em]{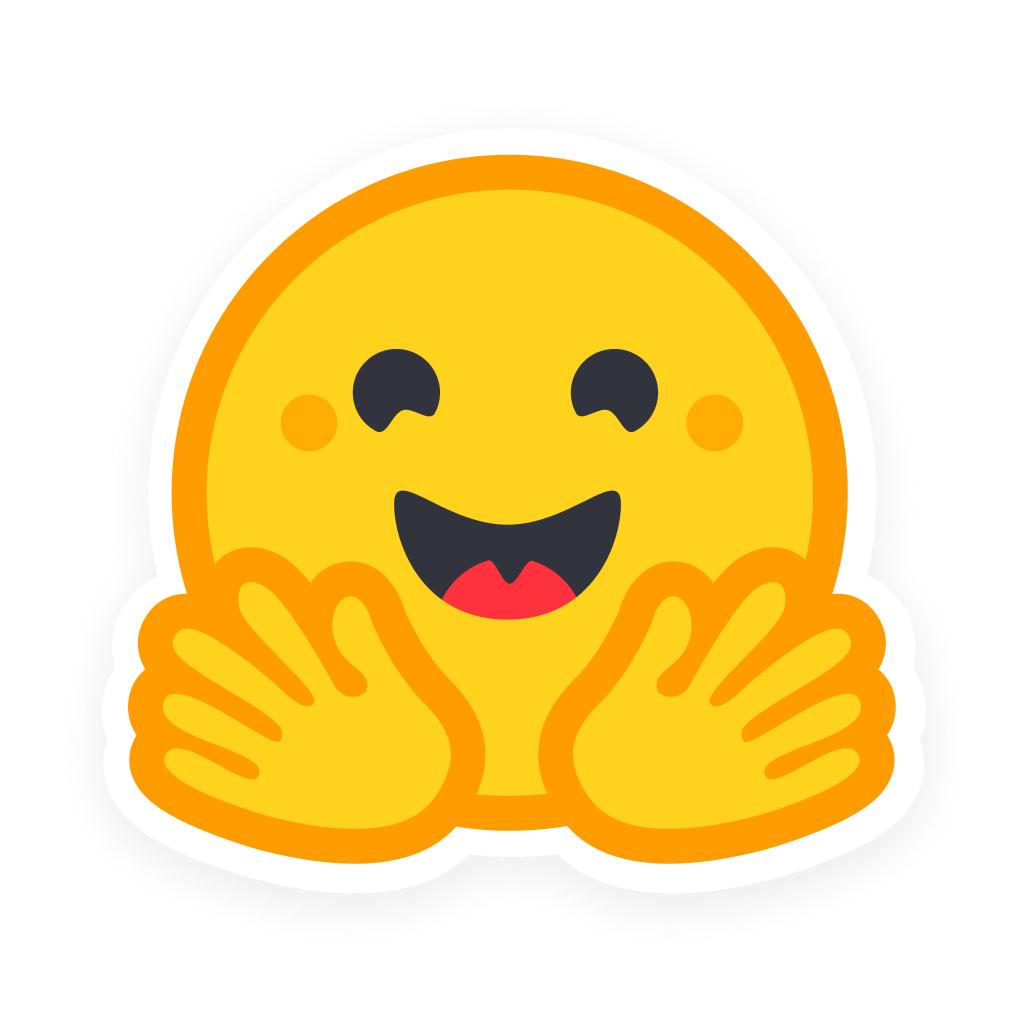}}%
\hspace{0.25em}\textbf{Model:}\enspace
\url{https://huggingface.co/collections/agentscope-ai/qwenpaw-flash}
\\[0.15em]
\raisebox{-0.22em}{%
  \includegraphics[height=1.15em]{figures/hf-logo.png}}%
\hspace{0.25em}\textbf{Dataset:}\enspace
\url{https://huggingface.co/datasets/agentscope-ai/Cowork-Gym}
\end{flushleft}
\end{abstract}

\begin{figure}[htbp]
  \centering
  \captionsetup{font=footnotesize,skip=4pt}
  \includegraphics[width=0.95\textwidth]{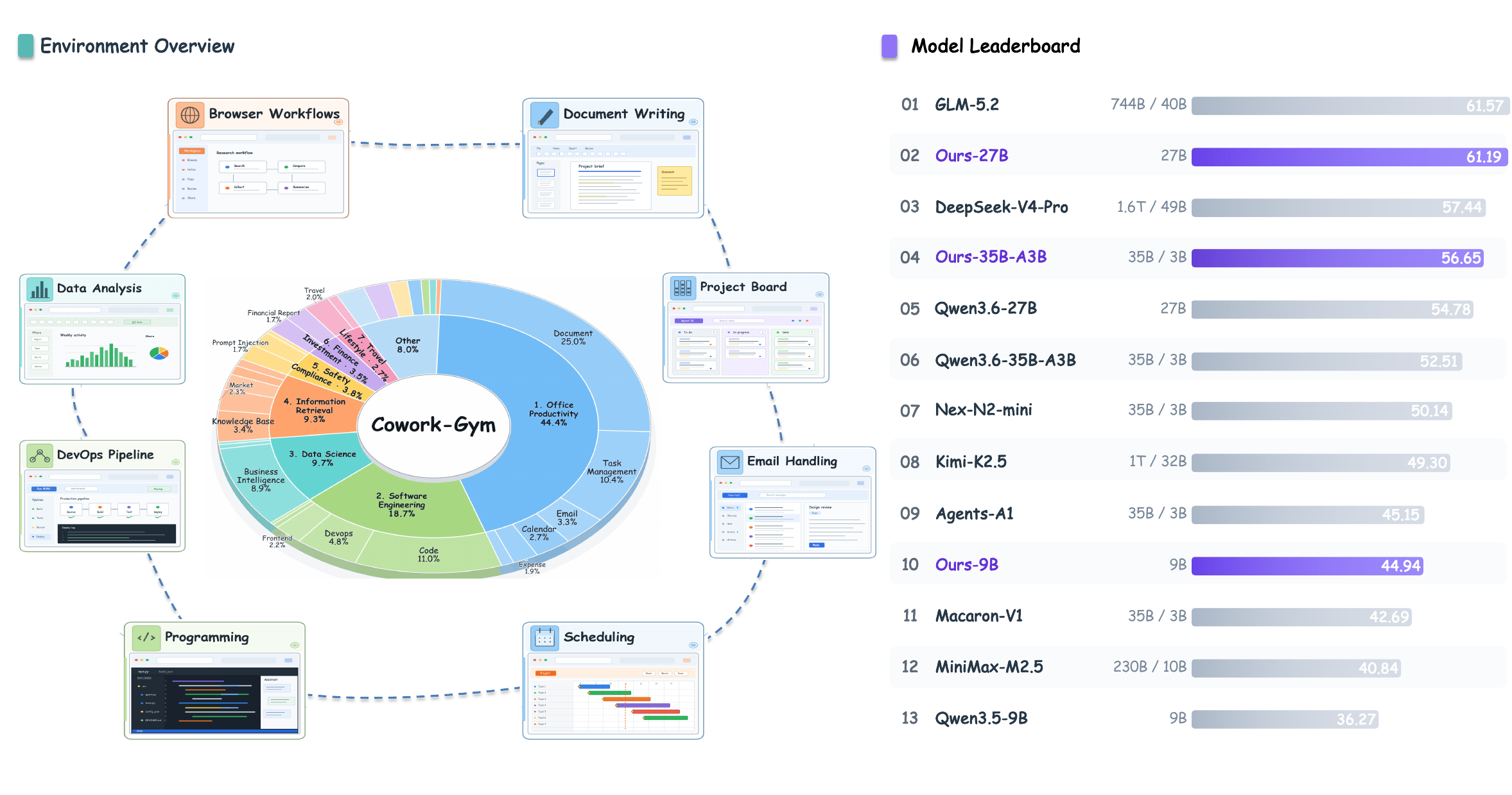}
  \caption{\textbf{Scale, diversity, and downstream impact of synthesized Cowork environments.}
  \textbf{Left:} Our environment-synthesis pipeline produces 64.9K tasks across 14 top-level scenarios, with representative workspaces illustrating diverse personal and professional workflows. \textbf{Right:} The leaderboard demonstrates the effectiveness of training on these environments: Ours-27B ranks second across six agentic benchmarks and achieves the highest average scores on PawBench and ClawEval among the evaluated models.}%
  \label{fig:environment-synthesis-overview}
\end{figure}

\section{Introduction}%
\label{sec:introduction}

Language-model agents increasingly undertake multi-step work that combines search, software development, computer use, and other capabilities \citep{webarena,osworld,agents_a1,glm5}, as illustrated by systems such as Claude Cowork, Codex, QwenPaw, and OpenClaw \citep{anthropic_cowork,openai_codex_cowork,qwenpaw,openclaw}. In these settings, an agent works through multiple tools in a persistent workspace, creating and revising artifacts as the task unfolds. At the same time, the user may supply new information, revise constraints, or replace task inputs, so the relevant state is shaped by both parties rather than fixed at the outset.

We use \emph{cowork} as a generic term for this interaction pattern, in which a user and an agent work toward an objective through a shared workspace. Progress is reflected in the artifacts they produce and the changes they make to that workspace. To train and evaluate this behavior, we define a \emph{Cowork environment} that specifies the task objective, initial workspace, available interfaces, user interaction policy, and verification criteria. It describes both the work to be completed and the conditions under which the interaction unfolds.

Constructing these environments at scale is challenging, which requires covering combinations of work domains, deliverable requirements, user profiles, and interaction patterns. Each combination should be realized in an executable workspace with compatible user actions and reliable outcome checks. These components must remain consistent as users reveal information, revise requirements, or replace inputs during execution. A final response may appear correct even when a required artifact is missing or the workspace is left in an incorrect state. Preparing and checking these conditions manually is costly; errors can make a task unsolvable or incorrectly scored. Recent advances in automatic environment synthesis \citep{scalable_environments,agent_world,clawenvkit,evoenv,verienv,cues} motivate us to explore automated synthesis for cowork environments, where workspace state and user requirements evolve during interaction.

We introduce \frameworkname, a scalable framework for synthesizing
Cowork environments (\cref{fig:method-overview}). It generates tasks
through two complementary routes: \emph{taxonomy-guided
generation} targets selected capabilities and scenario conditions,
while \emph{resource-grounded generation} derives tasks from existing
resources, such as code repositories and document collections.
Each task is instantiated as an executable environment and refined
through a \emph{build--execute--revise} loop. Sandbox execution and
verifier checks identify inconsistencies and guide repairs before
trajectory collection.

Within these environments, a \emph{stateful user simulator} responds
to the agent's progress through dialogue and direct workspace edits.
For example, after reviewing an initial analysis, it may replace
the source table, requiring the agent to revisit its inputs and
revise its outputs. This supports collaboration in which users
provide feedback and modify shared resources as work progresses.
We collect trajectories across agent harnesses and filter them
using artifact checks and execution records.

Using this framework, we synthesize 64.9K Cowork environments across 14 top-level scenarios, spanning office productivity, software engineering, data science, and other personal and professional workflows (\cref{fig:environment-synthesis-overview}). We use the collected trajectories and verification signals to post-train Qwen3.5-9B, Qwen3.6-27B, and Qwen3.6-35B-A3B through supervised fine-tuning, reinforcement learning, and optional on-policy distillation \citep{sdar,sdft}. Across six agentic benchmarks, the resulting models improve upon their base checkpoints in 17 of 18 matched model--benchmark comparisons, demonstrating consistent gains across model scales and task settings.

Our main contributions are threefold:
\begin{itemize}[leftmargin=1.3em]
  \item We develop \frameworkname for large-scale synthesis of cowork environments spanning diverse work domains and deliverables. It combines taxonomy-guided and resource-grounded task generation with a build--execute--revise loop to expand task coverage and validate environment consistency.
  \item We introduce a stateful user simulator that conditions conversational feedback and shared-workspace changes on the agent's progress, creating multi-turn collaboration in which the agent must adapt its ongoing work to revised requirements and inputs.
  \item We demonstrate the training value of synthesized environments across three model scales and six agentic benchmarks. Ours-27B ranks second overall among the evaluated models and achieves the best scores on PawBench and ClawEval, surpassing evaluated models with over 100B total parameters on both benchmarks.
\end{itemize}

\section{Problem Formulation}%
\label{sec:problem-formulation}

We consider Cowork tasks in which an agent uses multiple tools in a persistent workspace, while a stateful user may update the task through dialogue or workspace changes. To describe how such tasks become training data, we distinguish three levels: the environment specifies the task, a particular model--harness configuration realizes it as a trajectory, and filtering, selection, and transformation turn that trajectory into a training sample.

\begin{figure}[H]
  \centering
  \captionsetup{font=footnotesize,skip=4pt}
  \includegraphics[width=0.95\textwidth]{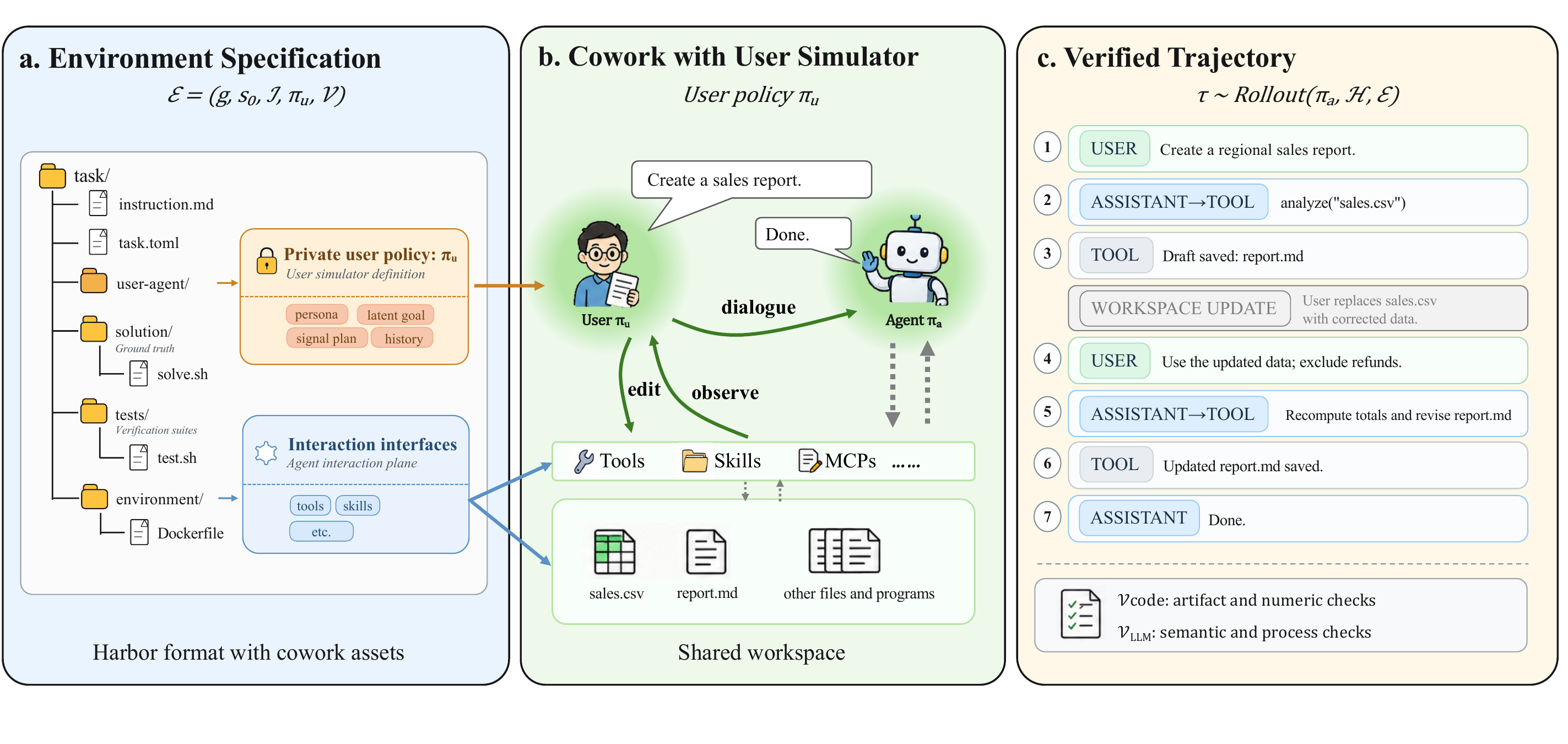}
  \caption{\textbf{Cowork environment packaging and rollout.}
  (a) Each environment is represented as a Harbor-compatible task package augmented with a user simulator, bundling the task instruction, runtime, reference solution, verifier, and user profiles.
  (b) During rollout, the user simulator communicates with the acting agent. It also observes and edits the shared workspace state in response to the agent's progress.
  (c) The resulting trajectory records user messages, agent actions, and tool observations; the hybrid verifier then evaluates it.
  }%
  \label{fig:harbor-format}
\end{figure}

\paragraph{Cowork environment.}
We represent a Cowork environment through five coupled components:
\begin{equation}
  \mathcal{E}=(g,s_0,\mathcal{I},\pi_U,\mathcal{V}),
  \label{eq:cowork-environment}
\end{equation}
Here, $g$ contains the task instruction and completion requirements, while $s_0$ contains the initial workspace and other persistent state. The interface set $\mathcal{I}$ covers the tools, skills, memory, permissions, and other ways to observe or modify that state, whereas $\pi_U$ specifies the stateful user's private context and intervention policy. Finally, the hybrid verifier $\mathcal{V}=(\mathcal{V}_{\mathrm{code}}, \mathcal{V}_{\mathrm{LLM}})$ combines deterministic checks of the final workspace and artifacts with LLM-based evaluation of semantic correctness and task-relevant process evidence.

\paragraph{Trajectory.}
Once an environment is defined, pairing it with an agent policy $\pi_A$ and harness $H$ yields a trajectory
\begin{equation}
  \tau \sim \operatorname{Rollout}(\pi_A,H,\mathcal{E}),
  \label{eq:rollout}
\end{equation}
where $\tau$ records the evolving workspace together with agent actions, user interventions, tool calls, and tool results. Because the harness mediates observations, tool schemas, execution, context management, and stopping behavior, the same environment may produce different trajectories under different policies or harnesses.

\paragraph{Training sample.}
Before a trajectory is used for training, a filtering operator $F(\tau;\mathcal{E})$ evaluates its task outcome and structural validity using evidence from $\mathcal{V}$ and checks of the event log. For each retained trajectory, $\operatorname{Format}_H$ then normalizes harness-specific messages, tool events, and metadata into the representation required by the downstream training stage, including supervision masks when applicable. The resulting dataset is
\begin{equation}
  \mathcal{D}_{\mathrm{train}}
  =\{\operatorname{Format}_H(\tau):F(\tau;\mathcal{E})=1\}.
  \label{eq:training-data}
\end{equation}


\section{Scalable Cowork Data Synthesis}%
\label{sec:environment-synthesis}
\label{sec:synthesis-framework}

\begin{figure}[!t]
  \centering
  \includegraphics[width=\textwidth]{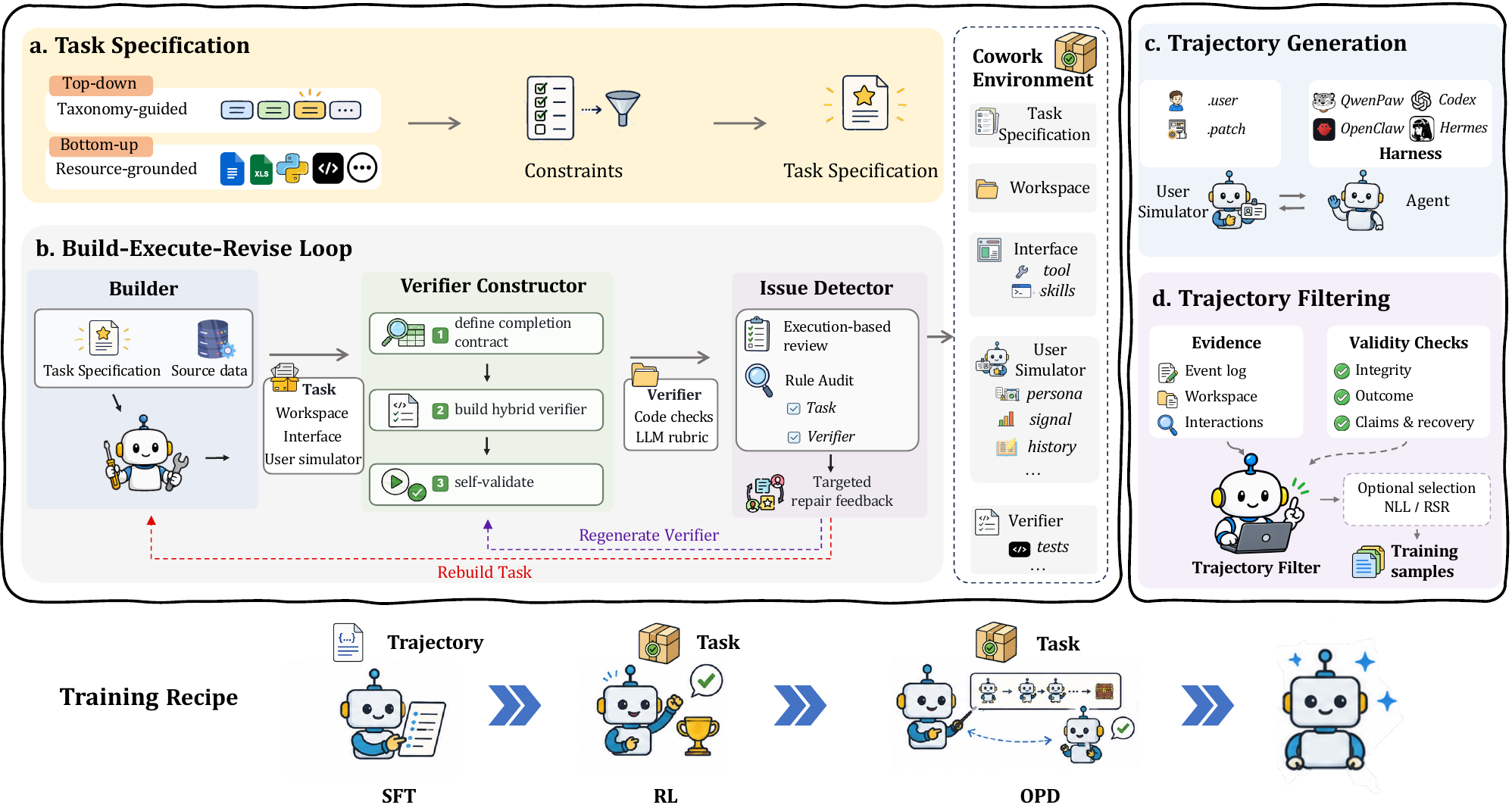}
  \caption{\textbf{Overview of Cowork environment synthesis and agent training.}
  \textbf{Left:} Environment synthesis starts with \textbf{(a)} \emph{Task Specification} (\cref{sec:task-seed-construction}) and proceeds through \textbf{(b)} the \emph{Build--Execute--Revise} loop (\cref{sec:environment-construction}).
  \textbf{Right:} \textbf{(c)} \emph{Trajectory Generation} with stateful user simulation is followed by \textbf{(d)} \emph{Filtering} (\cref{sec:trajectory-generation}).
  \textbf{Bottom:} The resulting environments and trajectories support agent training with SFT, RL, and OPD.}%
  \label{fig:method-overview}
\end{figure}

This section describes how \frameworkname constructs Cowork environments $\mathcal{E}$ and converts their rollouts into training samples (\cref{fig:method-overview}). We first derive the task specification $g$ from taxonomy conditions or grounded resources (\cref{sec:task-seed-construction}). We then instantiate and validate the components of $\mathcal{E}$ through a \emph{build--execute--revise} loop: the \emph{Builder} constructs $s_0$, $\mathcal{I}$, and $\pi_U$; the \emph{Verifier Constructor} constructs $\mathcal{V}$; and the \emph{Issue Detector} executes the assembled environment and routes defects for repair (\cref{sec:environment-construction}). During rollout, the \emph{stateful user simulator} implements $\pi_U$ while an acting policy $\pi_A$ operates through harness $H$ (\cref{sec:user-modeling}). Trajectory filtering and model-conditioned selection then convert the resulting trajectories $\tau$ into training samples (\cref{sec:trajectory-generation}).

\subsection{Task Specification}%
\label{sec:task-seed-construction}

We construct the task specification $g$ through two complementary routes. The \textbf{top-down, taxonomy-guided route} uses a task taxonomy to control coverage and composition. The \textbf{bottom-up, resource-grounded route} derives task requirements from existing work artifacts and resource collections based on their content and available operations. Regardless of the route, the resulting specification constrains the remaining components of $\mathcal{E}$ without prescribing a privileged solution trace.

\paragraph{Taxonomy-guided route.}
This route samples task conditions from a versioned, six-dimensional taxonomy covering capabilities, application scenarios, task structure and difficulty, user and deliverable requirements, input and output modalities, and safety. It includes 17 capability labels and 14 top-level application scenarios, with a three-level scenario hierarchy. Capabilities remain separate from application scenarios: a finance task, for example, may require data analysis without introducing a distinct \emph{financial data analysis} label. We discard incompatible combinations through schema and cross-field checks, then use accepted labels to condition generation, track coverage, and rebalance the task distribution. Capability definitions and the scenario hierarchy are summarized in \cref{app:task-taxonomy} of Appendix; the resulting distributions are reported in \cref{fig:task-scope-statistics}.

\paragraph{Resource-grounded route.}
This route starts from an existing resource and constructs a task around its content and available operations. It complements taxonomy-guided sampling with artifact-specific dependencies and interface constraints that are difficult to anticipate from labels alone. Illustrative sources include financial reports for cross-period analysis, legal documents for clause extraction and comparison, and diagnostic reports for summarizing recorded findings; repositories and paper collections similarly support software and literature tasks. We retain only tasks in which the resource contributes materially to completion. When applicable, complexity arises from cross-file dependencies, incomplete evidence, or required artifact transformations.

\subsection{Build--Execute--Revise Loop}%
\label{sec:environment-construction}

Given the task specification $g$, the \emph{Build--Execute--Revise} loop instantiates and validates the remaining components of $\mathcal{E}=(g,s_0,\mathcal{I},\pi_U,\mathcal{V})$. During \emph{Build}, the \emph{Builder} constructs $s_0$, $\mathcal{I}$, and $\pi_U$, while the \emph{Verifier Constructor} independently constructs $\mathcal{V}$ in an isolated sandbox. During \emph{Execute}, the \emph{Issue Detector} attempts to run the assembled environment and identifies defects across its components. During \emph{Revise}, each defect is returned to the responsible constructor, and the revised environment enters the next iteration. The loop continues until the environment satisfies the release criteria or reaches the maximum iteration budget.
We instantiate the \emph{Builder}, \emph{Verifier Constructor}, and
\emph{Issue Detector} as role-specialized agents on a common coding-agent
runtime with file, shell, web search and structured-output capabilities. Their prompts,
visible state, tool permissions, and sandboxes are configured separately.

\subsubsection{Builder}%
\label{sec:workspace-construction}


The \emph{Builder} compiles a task specification $g$ into an executable task package with three components: the initial workspace state $s_0$, the interface set $\mathcal{I}$, and the stateful user policy $\pi_U$. Their construction follows the \textbf{state visibility and isolation} principle. The acting agent receives only the public portion of the specification, the shared workspace state, and metadata for permitted interfaces. The user policy's private state, scheduled workspace updates, reference solutions, the implementation of $\mathcal{V}$, and construction metadata remain outside the agent's sandbox. Private information becomes observable only through a user message or an authorized workspace change produced by $\pi_U$.\footnote{The \emph{Builder} prompt is provided in Appendix~\ref{app:builder-prompt}.}

\paragraph{Initial workspace state $\boldsymbol{s_0}$.}
The initial workspace state contains the inputs, configurations, and editable files specified by $g$, together with the locations and formats of required deliverables. The \emph{Builder} records dependencies among these resources so that later changes remain consistent: replacing an input table may invalidate an analysis, while revising a source document may require regenerating its report. These dependencies also identify which artifacts the user policy may update and which outputs the verifier should inspect.


\paragraph{Interface set $\boldsymbol{\mathcal{I}}$.}
The interface set exposes only the operations required to complete the task, such as file operations, shell execution, application tools, skills, memory, or MCP services. Each interface specifies its version, working-directory rules, execution limits, and permission scope. Credentials, privileged operations, and implementation details remain outside the sandbox. This restricted boundary prevents unintended shortcuts and supports reproducible execution.


\paragraph{Stateful user policy $\boldsymbol{\pi_U}$.}
The \emph{Builder} creates the simulator's private profile, including its latent goal, knowledge boundary, signal plan, and permitted workspace operations. These assets determine what the simulated user knows, when it may provide additional information, and which workspace changes it may perform. They are not directly visible to the acting agent; their effects become observable only through simulator-generated messages or authorized updates to the shared workspace. \Cref{sec:user-modeling} describes how the \emph{stateful user simulator} executes $\pi_U$ during rollout.


\subsubsection{Verifier Constructor}%
\label{sec:verifier-construction}


The \emph{Verifier Constructor} builds a hybrid verifier $\mathcal{V}=(\mathcal{V}_{\mathrm{code}}, \mathcal{V}_{\mathrm{LLM}})$ in a sandbox isolated from the \emph{Builder}. Builder-owned artifacts remain read-only, and persistent writes are restricted to \texttt{tests/}. Each task requirement is mapped to a success condition, observable evidence, a checker or rubric item, and a weight. Expected values are derived from source data and the realized interaction, with reference materials used only for independent cross-checking.\footnote{The \emph{Verifier Constructor} prompt is provided in Appendix~\ref{app:verifier-constructor-prompt}.}


$\mathcal{V}_{\mathrm{code}}$ performs deterministic checks on the final workspace, artifacts, and trajectory, whereas $\mathcal{V}_{\mathrm{LLM}}$ evaluates semantic and procedural evidence. Together, they cover output validity, structural correctness, semantic cross-checking, and red-line constraints. Their results are combine  using normalized weights, with red-line failures limiting the maximum score. The complete verifier is self-validated on a reference completion and controlled failure cases before being passed to the \emph{Issue Detector}.

\subsubsection{Issue Detector}%
\label{sec:environment-validation}


The \emph{Issue Detector} reviews each assembled environment $\mathcal{E}$ before release. It combines execution-based review with a rule audit to assess consistency across all five components. Only environments that pass both stages proceed to trajectory generation.\footnote{The \emph{Issue Detector} prompt is provided in Appendix~\ref{app:issue-detector-prompt}.}

\paragraph{Execution-based review.}
The \emph{Issue Detector} runs the environment in a sandbox to inspect the task specification, initial workspace state, and interfaces, and to exercise the user policy's planned interventions. It tests the code-based and LLM-based verifier components on a valid reference completion and controlled failure states, checking whether they distinguish valid completion from incomplete, malformed, and untouched states.


\paragraph{Rule audit.}
The \emph{Rule Audit} applies fixed release criteria to the task package and execution evidence. It checks package structure, required fields, interface and permission allowlists, user assets, scheduled workspace patches, and leaks of answers, verifier logic, private information, or construction artifacts. Release also requires at least three meaningful score components and correct verifier behavior across the tested states.

\paragraph{Targeted repair feedback.}
Findings from both stages are routed by ownership: issues with the task specification, workspace state, interfaces, or user policy return to the \emph{Builder}, while verifier issues return to the \emph{Verifier Constructor}. Each revised environment is reviewed again in a reset sandbox until it passes or the iteration budget is exhausted, after which unresolved candidates are rejected.

\begin{figure}[!t]
  \centering
  \includegraphics[width=\textwidth]{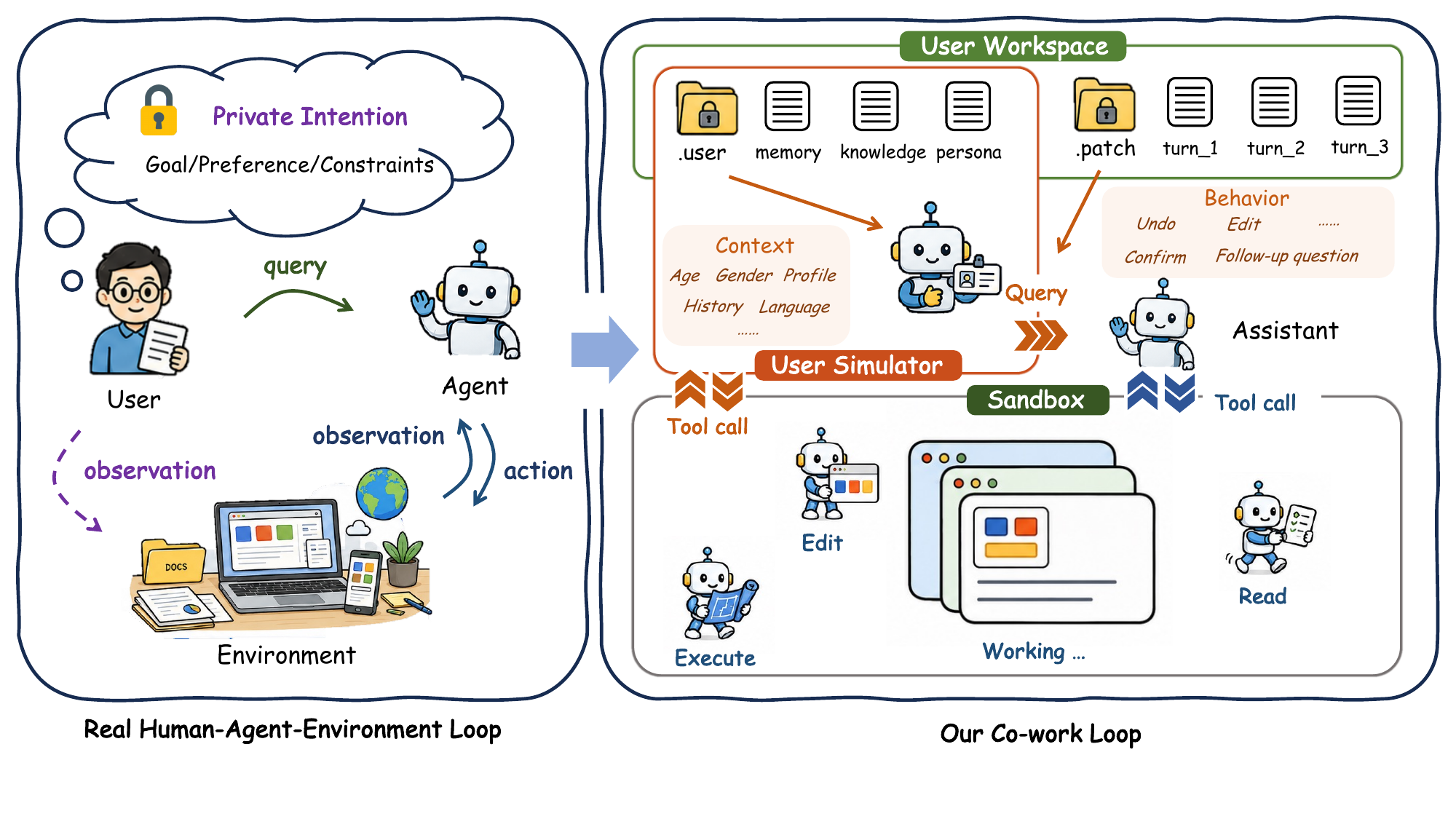}
  \caption{\textbf{Cowork environment synthesis and stateful user simulation.}
  \textbf{Left:} The interaction loop among user, agent, and the environment in the real world.
  \textbf{Right:} The \emph{user simulator} runs outside the acting-agent sandbox and conditions each intervention on private user state and current task progress. The \texttt{.user/} directory stores persistent context; \texttt{.patch/} provides turn-specific behavioral instructions and authorized workspace updates. The acting agent observes only public messages and the resulting changes(\cref{sec:user-modeling}).}%
  \label{fig:user-simulator}
\end{figure}

\subsection{Stateful User Simulator}%
\label{sec:user-modeling}

As shown on the right of \cref{fig:user-simulator}, the \emph{stateful user simulator} is the executable implementation of the user policy $\pi_U$. The \emph{Builder} initializes its private state and permissions during environment construction (\cref{sec:workspace-construction}); during rollout, the simulator runs outside the acting agent's sandbox and reacts to intermediate results, corrects misunderstandings, reveals new information, and modifies the shared workspace. The acting agent observes only public messages and resulting workspace changes.

\paragraph{Private user state.}
Each environment defines a private user profile containing a persona, latent goal, knowledge boundary, interaction history, and signal plan. The persona governs how the user communicates, while the knowledge boundary limits the simulator to information that the represented user could reasonably know. The signal plan identifies possible stages of interaction without fixing messages in advance. The initial query reveals enough information to begin the task, while further constraints or preferences are disclosed only when relevant.

\paragraph{State-dependent interaction.}
The user simulator communicates with the acting agent to finish the task. After each agent work segment, the simulator reviews the response, public interaction history, and permitted workspace state. It may intervene when the task remains ambiguous, a draft needs revision, an error occurs, or an input changes. Because each response and workspace update depends on the agent's actual progress, the same initial query can produce different follow-ups across rollouts.

\paragraph{Workspace intervention.}
In Cowork settings, users and agents collaboratively create and revise artifacts in a shared workspace. In addition to conversational feedback, the simulator may revise an agent-produced draft or apply controlled file patches or tool-based edits to update source data before the next user message. Such changes can invalidate earlier observations and intermediate results, requiring the agent to re-read relevant files, revise assumptions, and update dependent artifacts.

\subsection{Trajectory Generation and Filtering}%
\label{sec:trajectory-generation}
\label{sec:trajectory-filtering}


For each released environment $\mathcal{E}=(g,s_0,\mathcal{I},\pi_U,\mathcal{V})$, the acting agent starts from $s_0$ and operates through $\mathcal{I}$, while the \emph{stateful user simulator} executes $\pi_U$ outside the agent's sandbox. The resulting trajectory $\tau$ includes an event log  and final workspace state. The filtering operator $F(\tau;\mathcal{E})$ determines whether $\tau$ provides reliable supervision before model-conditioned selection is applied.

\paragraph{Cross-harness rollout with user simulator.}
To enable comparable rollouts across harnesses, we represent each $\mathcal{E}$ using the Harbor task format \citep{harbor}, as illustrated in \cref{fig:harbor-format}. This representation keeps $\mathcal{E}$ fixed while allowing the acting policy $\pi_A$ or harness $H$ to vary. The acting agent and \emph{stateful user simulator} interact through public messages and workspace changes.
We collect training trajectories with QwenPaw and OpenClaw \citep{qwenpaw,openclaw}, and evaluate transfer to Hermes Agent \citep{hermes_agent} (\cref{sec:cross-harness-results}).

\paragraph{Trajectory filtering.} The filtering operator $F(\tau;\mathcal{E})$ determines whether an individual trajectory provides reliable supervision along three dimensions:
\begin{enumerate}[label=\textbf{(\arabic*)},leftmargin=*,nosep]
  \item \textbf{Record integrity.} The event log must be complete, uncorrupted, and free of leaked private information.
  \item \textbf{Task outcome.} The hybrid verifier evaluates the final workspace state, produced artifacts, and task-relevant interaction evidence.
  \item \textbf{Claims and recovery.} We compare success claims against observed outcomes and reject unsupported claims or unresolved failures. Intermediate tool errors are permitted when the agent recovers and reaches a verified outcome.
\end{enumerate}

\paragraph{Model-conditioned selection.}
The preceding filtering stage identifies valid trajectories that can be used for training. Beyond this validity-based filtering, we investigate whether model-conditioned metrics can further improve training effectiveness by prioritizing trajectories according to their potential learning value for a target student $\pi_\theta$. We use average negative log-likelihood (NLL) and Rank--Surprisal Ratio (RSR) \citep{yang-etal-2026-reasoning} as alternative criteria under a fixed retention budget. Let $\mathcal{M}_\tau$ denote the scored token positions in trajectory $\tau$, with $y_t$ the recorded token and $h_t$ its preceding interaction context. The two scores are
\begin{equation}
  \begin{aligned}
    \mathrm{NLL}_\theta(\tau)
    &=-\frac{1}{|\mathcal{M}_\tau|}
      \sum_{t\in\mathcal{M}_\tau}\log\pi_\theta(y_t\mid h_t),\\
    \mathrm{RSR}_\theta(\tau)
    &=\frac{\sum_{t\in\mathcal{M}_\tau}\min(r_{\theta,t},r_{\max})}
      {|\mathcal{M}_\tau|\,\mathrm{NLL}_\theta(\tau)}.
  \end{aligned}
  \label{eq:model-conditioned-scores}
\end{equation}
Here, $r_{\theta,t}$ is the rank of $y_t$ under $\pi_\theta(\cdot\mid h_t)$, with rank one assigned to the most probable token, and $r_{\max}$ caps the rank. NLL measures how unexpected the scored tokens are to the student. RSR divides the mean clipped rank by NLL, combining relative token preference and surprisal as a proxy for data--student suitability.

To limit shifts in the trajectory-length distribution, we rank scores within equal-frequency length bins. For a retention fraction $0<\rho\leq 1$, NLL retains the central fraction $\rho$ of trajectories in each bin, whereas RSR retains the fraction $\rho$ with the lowest scores (see \cref{sec:trajectory-filtering-results} for the experimental settings).
Finally, $\operatorname{Format}_H$ normalizes the selected harness-specific trajectories into the representation required by the downstream training stage, producing samples for $\mathcal{D}_{\mathrm{train}}$.
\section{Agentic Post-Training}%
\label{sec:post-training}

We consider two post-training routes. The first directly adapts the model through SFT followed by outcome-based RL\@. The second uses on-policy distillation: a student initialized from the base checkpoint learns the agentic capabilities of a same-size SFT or SFT+RL teacher from interaction histories generated by the student itself. Direct adaptation can improve agentic performance while weakening capabilities outside the training distribution \citep{sdft,general_capability_forgetting}. Motivated by work on on-policy distillation \citep{sdft,sdar}, we investigate whether the second route improves agent performance while better preserving the model's general capabilities.

\paragraph{Supervised fine-tuning.}
We perform SFT on filtered trajectories converted from harness-specific formats into the target model's training format. This conversion standardizes the representation while preserving the recorded actions and observations. The loss is computed only on tokens generated by the agent, including assistant messages and tool calls. Public user messages, tool observations, and workspace events are included as context but excluded from the loss; private simulator state and host metadata are excluded from the training input. The mixture includes successful executions, revision-driven interactions, and grounded recoveries from tool errors.

\paragraph{Outcome-based reinforcement learning.}
Starting from the SFT checkpoint, we apply Group Relative Policy Optimization (GRPO) using the environment's verification suite $\mathcal{V}=(\mathcal{V}_{\mathrm{code}},\mathcal{V}_{\mathrm{LLM}})$ as outcome supervision. For each environment, we sample a group of trajectories and denote the scores assigned to $\tau_i$ by the two verifier components as $s_i:=\mathcal{V}_{\mathrm{code}}(\tau_i)\in[0,1]$ and $c_i:=\mathcal{V}_{\mathrm{LLM}}(\tau_i)\in[0,1]$. The code-based score checks required artifacts and deterministic constraints, while the LLM-based score applies a reference-conditioned rubric to semantic correctness, coverage, contradictions, and other task-specific criteria. We combine them as
\begin{equation}
  R_{\mathcal{E}}(\tau_i)=
  \begin{cases}
    s_i^{\frac{w_s}{w_s+w_c}}
    c_i^{\frac{w_c}{w_s+w_c}}, & s_i>0 \text{ and } c_i>0,\\
    0, & \text{otherwise},
  \end{cases}
  \label{eq:rl-outcome-reward}
\end{equation}
where $w_s,w_c>0$ determine the relative importance of executable verification and semantic correctness. Their normalized exponents lie in $(0,1)$ and sum to one, so only the ratio $w_s:w_c$ affects the reward. The weighted geometric mean favors trajectories that score well on both criteria and assigns zero reward whenever either score is zero. Compared with an arithmetic mean using the same normalized weights, it more strongly penalizes trajectories with an imbalanced score pair. We use standard GRPO updates and exclude groups with zero reward variance, since equal rewards provide no within-group reward differences for estimating relative advantages. For MoE models, we freeze the router weights during training \citep{blakeman2025nemotron}.

\paragraph{On-policy distillation.}
We initialize the student from the base checkpoint used before agentic post-training and use a same-size SFT or SFT+RL checkpoint as the teacher. The student interacts with the environment to collect trajectories. At each agent-generated token position, the teacher provides a next-token probability distribution conditioned on the same interaction prefix. We train the student to match this distribution by minimizing $D_{\mathrm{KL}}(\pi_T\,\Vert\,\pi_\theta)$ at these positions. User messages, tool observations, and workspace events supply context but are not prediction targets. This provides teacher supervision on interaction histories encountered by the student itself.


\section{Experiments}%
\label{sec:experiments}

Our experiments examine the distribution and complexity of the synthesized data, its downstream training value, and the trade-off between agentic gains and general-capability retention through four questions:
\begin{enumerate}[label=\textbf{Q\arabic*.},leftmargin=*,itemsep=0.2em,topsep=0.3em]
  \item How are the synthesized tasks distributed across work scenarios and capabilities, and what interaction and execution demands do they place on agents? (\cref{sec:environment-statistics})
  \item Does training on the synthesized data yield agentic gains across model scales and evaluation benchmarks? (\cref{sec:main-results})
  \item How do environment exposure, multi-turn supervision, harness diversity, and trajectory selection affect the training value and transferability of the synthesized data? (\cref{sec:design-analysis})
  \item Can on-policy distillation transfer agentic gains while reducing the loss of general capabilities relative to direct SFT? (\cref{sec:opd-generalization-results})
\end{enumerate}

\subsection{Experimental Setup}%
\label{sec:experimental-setup}

We evaluate models on six benchmarks: Workspace-Bench \citep{workspace_bench}, ALE (Agents' Last Exam) \citep{agents_last_exam}, WildClawBench \citep{wildclawbench}, PawBench \citep{pawbench}, ClawEval \citep{claw_eval}, and ClawMark \citep{clawmark}. These benchmarks span file-centric workflows, professional tasks, long-horizon tool use, multimodal inputs, and dynamic multi-turn interaction. We use each benchmark's official tasks, metrics, and evaluators. OpenClaw serves as the default harness unless another runtime is required; the main results report PawBench under OpenClaw, with QwenPaw \citep{qwenpaw} and Hermes Agent \citep{hermes_agent} used only for cross-harness analysis.

Our base checkpoints are Qwen3.5-9B, Qwen3.6-27B, and Qwen3.6-35B-A3B \citep{qwen35_9b,qwen36_27b,qwen36_35b}. The comparison models are MiniMax-M2.5, GLM-5.2, Kimi-K2.5, DeepSeek-V4-Pro, Agents-A1, Nex-N2-mini, and Macaron-V1-Tall \citep{minimax_m25,glm52,kimi_k25,deepseek_v4,agents_a1,nex_n2_mini,macaron_v1}.


\begin{table}[!tbp]
  \centering
  \includegraphics[width=\textwidth]{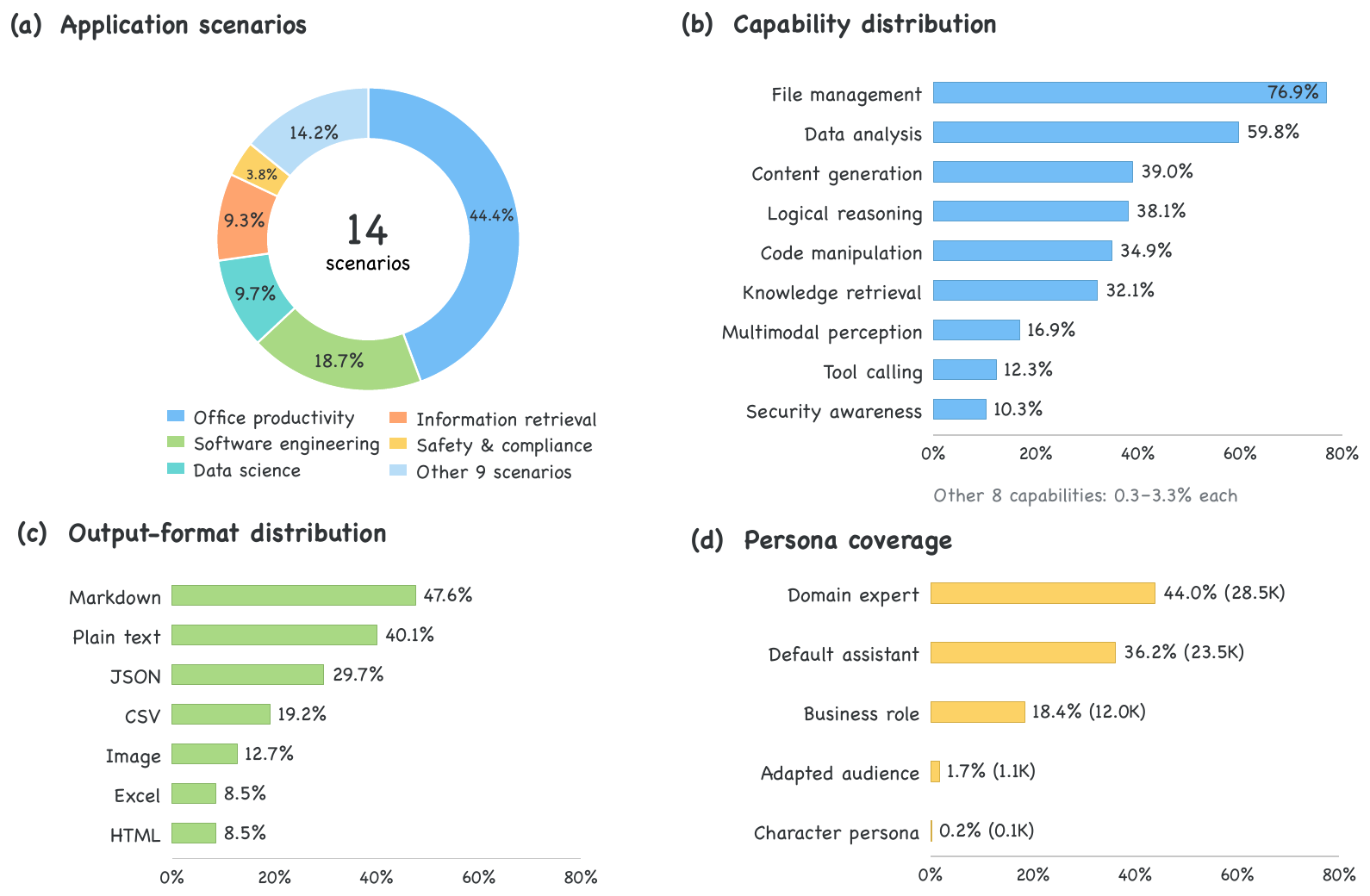}
  \captionof{figure}{Distribution across \textbf{(a)} application scenarios, \textbf{(b)} capability labels, \textbf{(c)} output formats, and \textbf{(d)} persona categories. Percentages are computed over 64,888 environments.}%
  \label{fig:task-scope-statistics}
  \vspace{0.4em}
  \begin{minipage}{\linewidth}
    \centering
    \captionof{table}{\textbf{Trajectory scale and interaction structure.}
    Panel (a) summarizes SFT trajectory filtering, RL environment scale,
    and execution depth after judge filtering;
    panel (b) describes interaction composition in the raw corpus.}
    \label{tab:trajectory-structure}
    \small
    \setlength{\tabcolsep}{3pt}
    \renewcommand{\arraystretch}{1.15}
    \medskip
    \begin{tabularx}{\linewidth}{@{}>{\hsize=1.22\hsize\linewidth=\hsize\raggedright\arraybackslash}Xr@{}p{0.02\linewidth}@{}>{\hsize=0.78\hsize\linewidth=\hsize\raggedright\arraybackslash}Xr@{}}
        \multicolumn{2}{@{}l}{\textbf{(a) Collection and filtering}}
        & & \multicolumn{2}{@{}l}{\textbf{(b) Interaction composition}} \\
        \cmidrule[\heavyrulewidth]{1-2}\cmidrule[\heavyrulewidth]{4-5}
        \textbf{Measure} & \textbf{Statistic}
        & & \textbf{Measure} & \textbf{Percentage} \\
        \cmidrule{1-2}\cmidrule{4-5}
        \multicolumn{2}{@{}l}{\textit{Training data}}
        & & \multicolumn{2}{@{}l}{\textit{Interaction types}} \\
        Raw trajectories & 77,729
        & & Single-turn trajectories & 67.01\% \\
        After validity-based filtering & 24,732
        & & Multi-turn trajectories & 32.99\% \\
        After model-conditioned selection (SFT trained) & 17,312
        & & & \\
        \addlinespace[3pt]
        RL training environments & 2,873
        & & & \\
        \cmidrule{1-2}\cmidrule{4-5}
        \multicolumn{2}{@{}l}{\textit{Trajectories after validity-based filtering}}
        & & \multicolumn{2}{@{}l}{\textit{User turns (multi-turn)}} \\
        Tool-using steps, median & 10
        & & 2--4 turns & 63.16\% \\
        ${}>20$ tool-using steps & 3,437 (13.90\%)
        & & 5--8 turns & 22.53\% \\
        ${}>40$ tool-using steps & 303 (1.23\%)
        & & $\geq 9$ turns & 14.31\% \\
        \cmidrule[\heavyrulewidth]{1-2}\cmidrule[\heavyrulewidth]{4-5}
    \end{tabularx}
    \par\smallskip
    \begin{minipage}{\linewidth}
        \scriptsize
        \textit{Note.}
        Model-conditioned selection uses NLL/RSR\@. A tool-using step is an assistant
        message containing at least one tool call. Single- and multi-turn denote
        trajectories with one and more than one user-role message, respectively.
    \end{minipage}
  \end{minipage}
\end{table}

\subsection{Data Scale and Diversity}%
\label{sec:environment-statistics}

We characterize the synthesized data from the environment and trajectory aspects. \Cref{fig:task-scope-statistics} reports the scale and composition of the environment pool, while \cref{tab:trajectory-structure} summarizes the interaction and execution structure of the collected trajectories.

\paragraph{Environment coverage.}
The pipeline synthesizes 64,888 environments spanning 14 application scenarios and 17 capabilities, with variation across scenarios, capabilities, and deliverables (\cref{fig:task-scope-statistics}). Scenario coverage is broad but non-uniform: office productivity and software engineering form the largest groups, while less frequent domains remain represented. These statistics capture the scale and compositional breadth of the environment pool, supporting the pipeline's capacity to synthesize heterogeneous task configurations without implying uniform coverage across labels.

\noindent\textbf{Trajectory structure.}
Table~\ref{tab:trajectory-structure} summarizes the corpus scale and execution distributions.
Validity-based filtering first retains 24,732 of 77,729 raw trajectories (31.82\%),
after which model-conditioned selection retains 17,312 trajectories for training.
The evidence-valid set contains a median of 10 tool-using assistant steps per trajectory,
with the middle 50\% spanning 6--16 steps, compared with 6--19 steps in the raw corpus.
Single- and multi-turn trajectories account for 67.01\% and 32.99\% of the raw corpus, respectively.
Single-turn interactions can involve substantial tool execution:
their median step count is 10, with an interquartile interval of 6--17 steps,
while multi-turn trajectories have a median of 15 and an interval of 9--24 steps.
The difference also extends into the upper tail:
32.69\% of multi-turn trajectories exceed 20 tool-using steps,
compared with 15.80\% of single-turn trajectories.
These statistics capture complementary aspects of interaction structure:
the number of user turns and the depth of tool execution within a trajectory.

\subsection{Main Results}%
\label{sec:main-results}

\begin{table}[!tbp]
  \centering
  \newcommand{\mainbest}[1]{\textbf{#1}}
  \newcommand{\mainsecond}[1]{\underline{#1}}
  \makebox[\textwidth][c]{%
  \begin{threeparttable}
    \caption{Main results on six agentic benchmarks under \textsc{OpenClaw}. Entries report \emph{average score\,/\,pass rate} where both are available and average score otherwise; Avg.\ denotes the mean score across the six benchmarks, and MoE size is total / activated parameters. The best and second-best scores in each metric column are \textbf{bolded} and \underline{underlined}, respectively.}%
    \label{tab:main-results}
    \scriptsize
    \setlength{\tabcolsep}{2.8pt}
    \renewcommand{\arraystretch}{1.26}
    \begin{tabular}{@{}lcccccccc@{}}
      \toprule
      Model &
      Params. &
      Avg. &
      \makecell{Workspace\\Bench} &
      ALE &
      \makecell{WildClaw\\Bench} &
      PawBench &
      ClawEval &
      ClawMark \\
      \midrule
      \multicolumn{9}{@{}l}{\textit{$>$100B models}} \\[-0.35em]
      \cmidrule{1-9}
      MiniMax-M2.5      & 230B / 10B & 0.4084 & 0.2517 & 0.2622 / 9.90\% & 0.3936 & 0.5431 / 10.75\% & 0.6526 / 59.60\% & 0.3472 \\
      GLM-5.2           & 744B / 40B & \mainbest{0.6157} & 0.5178 & \mainbest{0.3040} / \mainbest{12.80}\% & \mainbest{0.5866} & 0.7878 / 20.70\% & 0.7601 / \mainsecond{80.70}\% & \mainbest{0.7377} \\
      Kimi-K2.5         & 1T / 32B & 0.4930 & 0.3949 & 0.2880 / 11.30\% & 0.3486 & 0.7331 / 13.44\% & 0.6967 / 69.6\% & 0.4968 \\
      DeepSeek-V4-Pro   & 1.6T / 49B & 0.5744 & \mainbest{0.5525} & 0.2760 / \mainsecond{12.40}\% & \mainsecond{0.4614} & 0.7729 / 15.32\% & 0.7727 / 75.5\% & 0.6109 \\
      \addlinespace[0.25em]
      \multicolumn{9}{@{}l}{\textit{9B models}} \\[-0.35em]
      \cmidrule{1-9}
      Qwen3.5-9B & 9B & 0.3627 & 0.2372 & 0.0205 / 0.70\% & 0.2302 & 0.7071 / 24.55\% & 0.6535 / 62.10\% & 0.3276 \\
      \rowcolor{copawblue!5}
      \textbf{Ours-9B} & 9B & 0.4494 & 0.3836 & 0.1544 / 4.30\% & 0.2438 & 0.7699 / 31.82\% & 0.7259 / 75.80\% & 0.4185 \\
      \addlinespace[0.25em]
      \multicolumn{9}{@{}l}{\textit{35B-A3B models}} \\[-0.35em]
      \cmidrule{1-9}
      Qwen3.6-35B-A3B & 35B / 3B & 0.5251 & 0.5077 & 0.1794 / 5.00\% & 0.4300 & 0.7739 / 30.91\% & 0.7254 / 67.70\% & 0.5344 \\
      Agents-A1 & 35B / 3B & 0.4515 & 0.3599 & 0.2263 / 9.20\% & 0.2794 & 0.7045 / 25.3\% & 0.7331 / 78.30\% & 0.4057 \\
      Nex-N2-mini & 35B / 3B & 0.5014 & 0.4426 & 0.2112 / 7.10\% & 0.3201 & 0.6878 / 21.77\% & 0.7389 / 78.90\% & 0.6078 \\
      Macaron-V1 & 35B / 3B & 0.4269 & 0.2562 & 0.1617 / 7.10\% & 0.2721 & 0.7120 / 18.55\% & 0.6796 / 70.20\% & 0.4796 \\
      \rowcolor{copawblue!5}
      \textbf{Ours-35B-A3B} & 35B / 3B & 0.5665 & 0.5227 & 0.2584 / 8.50\% & 0.4248 & \mainsecond{0.8187} / 37.27\% & \mainsecond{0.7734} / \mainbest{82.00}\% & 0.6009 \\
      \addlinespace[0.25em]
      \multicolumn{9}{@{}l}{\textit{27B models}} \\[-0.35em]
      \cmidrule{1-9}
      Qwen3.6-27B & 27B & 0.5478 & 0.5230 & 0.2311 / 8.50\% & 0.3725 & 0.8057 / \mainsecond{39.09}\% & 0.7311 / 68.90\% & 0.6233 \\
      \rowcolor{copawblue!5}
      \textbf{Ours-27B} & 27B & \mainsecond{0.6119} & \mainsecond{0.5409} & \mainsecond{0.2903} / 10.6\% & 0.4241 & \mainbest{0.8400} / \textbf{41.82}\% & \mainbest{0.8508} / 80.6\% & \mainsecond{0.7250} \\
      \bottomrule
    \end{tabular}
  \end{threeparttable}
  }%
\end{table}

Our models in \cref{tab:main-results} are first trained with SFT on 17,312 trajectories selected from the 24,732 evidence-valid trajectories retained from an initial collection of 77,729. Starting from the SFT checkpoints, we then perform outcome-based RL on 2,873 environments.
Averaged over the six benchmarks, the 9B, 27B, and 35B-A3B models improve over their matched baselines by \textbf{+8.7}, \textbf{+6.4}, and \textbf{+4.1 points}, respectively, with gains in 17 of the 18 model--benchmark comparisons. Beyond this broad improvement, the gains exhibit distinct patterns across backbones: 9B benefits most on Workspace-Bench (+14.64 points) and ALE (+13.39), whereas 27B improves most on ClawEval (+11.97) and ClawMark (+10.17). This suggests that the same training collection helps address different weaknesses across models, rather than yielding gains that simply increase with parameter count.

Ours-27B ranks second by the six-benchmark mean (0.6119 versus 0.6157 for GLM-5.2), while achieving the best scores on PawBench and ClawEval among all evaluated models, including those with over 100B total parameters. Within the 35B-A3B group, Ours-35B-A3B also achieves the highest mean score. The gains nevertheless have a clear boundary: on WildClawBench, 9B improves by only 1.36 points and 35B-A3B declines by 0.52 points, while all three of our models remain substantially behind GLM-5.2.

\subsection{Ablation Studies}%
\label{sec:design-analysis}

To better understand the factors underlying the observed performance gains, we next examine four complementary dimensions of the data pipeline: how \textbf{supervised environment exposure} interacts with model capacity; the value of \textbf{interactive supervision} from multi-turn trajectories; \textbf{cross-harness transfer} across execution runtimes; and \textbf{trajectory selection} via model-conditioned filtering.

\subsubsection{Supervised Environment Exposure}%
\label{sec:environment-scaling-results}

\begin{figure}[H]
  \centering
  \includegraphics[width=\textwidth]{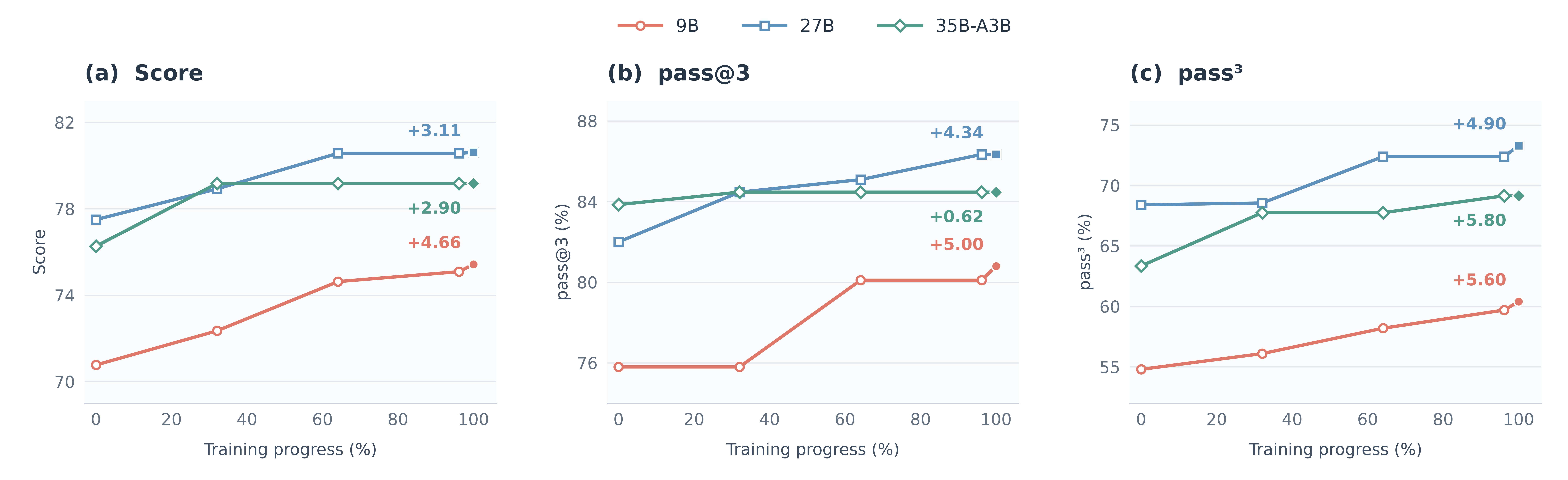}
  \caption{ClawEval performance of Ours-9B, Ours-27B, and Ours-35B-A3B over supervised training, with progress normalized to the final checkpoint.}%
  \label{fig:sft-environment-exposure}
\end{figure}

Figure~\ref{fig:sft-environment-exposure} tracks ClawEval performance from the initial to the final supervised-training checkpoint. Comparing these endpoints, all three model variants improve across all three metrics. The 9B model achieves the largest gains in Score (+4.66) and pass@3 (+5.00 percentage points), showing that the smaller dense model continues to acquire agentic capabilities throughout training. The 27B model also improves at intermediate checkpoints and achieves the highest final performance across all three metrics. The 35B-A3B model exhibits a more metric-dependent trajectory: Score and pass@3 plateau relatively early, whereas $\text{pass}^3$ continues to improve and yields the largest endpoint gain in this metric (+5.80 percentage points). These patterns indicate that the benefits of continued supervised training accrue at different rates across model scales and metrics; a plateau in one metric does not necessarily imply that further training provides no benefit on others.

\subsubsection{Interactive Supervision}%
\label{sec:user-simulation-results}
\leavevmode\par
\vspace{-0.5em}

\begin{table}[H]
  \centering
  \begin{threeparttable}
    \caption{Effect of interactive supervision on 27B models. Metrics are reported on a 0--100 scale; green badges indicate absolute gains over Qwen3.6-27B.}%
    \label{tab:user-simulation-v2}
    \small
    \setlength{\tabcolsep}{3pt}
    \renewcommand{\arraystretch}{1.22}
    \begin{tabularx}{\textwidth}{lcc*{5}{>{\centering\arraybackslash}X}}
      \toprule
      \multirow{2}{*}{\textbf{Model}} &
      \multirow{2}{*}{\makecell{\textbf{Single}\\[-1pt]\textbf{turn}}} &
      \multirow{2}{*}{\makecell{\textbf{Multi}\\[-1pt]\textbf{turn}}} &
      \multicolumn{2}{c}{\cellcolor{copawgroupbg}\textbf{ClawEval}} &
      \multicolumn{2}{c}{\cellcolor{copawgroupbg}\textbf{ClawEval-MT}} &
      \multicolumn{1}{c}{\cellcolor{copawgroupbg}\textbf{ClawMark}} \\
      & & & \makebox[\linewidth][c]{\textbf{Score}}
        & \makebox[\linewidth][c]{\textbf{Pass@3 (\%)}}
        & \makebox[\linewidth][c]{\textbf{Score}}
        & \makebox[\linewidth][c]{\textbf{Pass (\%)}}
        & \makebox[\linewidth][c]{\textbf{Score}} \\
      \midrule
      \rowcolor{copawgroupbg!45}
      Qwen3.6-27B
        & -- & --
        & 73.11 & 68.9 & 80.57 & 71.1 & 62.33 \\

      \addlinespace[3pt]
      Ours-27B
        & \checkmark & --
        & \mbox{84.13\,\gainbadge{11.02}}
        & \mbox{79.7\,\gainbadge{10.8}}
        & \mbox{84.01\,\gainbadge{3.44}}
        & \mbox{83.3\,\gainbadge{12.2}}
        & \mbox{69.27\,\gainbadge{6.94}} \\


      \addlinespace[3pt]
      \rowcolor{copawblue!5}
      \textbf{Ours-27B}
        & \checkmark & \checkmark
        & \mbox{85.08\,\gainbadge{11.97}}
        & \mbox{80.6\,\gainbadge{11.7}}
        & \mbox{84.44\,\gainbadge{3.87}}
        & \mbox{81.6\,\gainbadge{10.5}}
        & \mbox{72.50\,\gainbadge{10.17}} \\

      \bottomrule
    \end{tabularx}
  \end{threeparttable}
\end{table}

Table~\ref{tab:user-simulation-v2} shows that adding multi-turn trajectories to the single-turn training set improves average scores across all three benchmarks, with the largest gain on ClawMark (\textbf{+3.23 points}). This suggests that interactive supervision provides complementary benefits for cowork task performance.

\subsubsection{Cross-Harness Transfer}%
\label{sec:cross-harness-results}
\leavevmode\par
\vspace{-0.5em}

\begin{table}[H]
  \centering
  \caption{Cross-harness generalization of Qwen3.5-9B on PawBench. Results are pass rates; green badges show absolute gains over the base model, with darker shades indicating larger gains.}%
  \label{tab:cross-harness}
  \small
  \setlength{\tabcolsep}{10pt}
  \renewcommand{\arraystretch}{1.22}
  \begin{tabularx}{\textwidth}{l*{3}{>{\centering\arraybackslash}X}}
    \toprule
    \rowcolor{copawgroupbg}
    \textbf{Training source}
      & \textbf{QwenPaw}\enspace\seentag
      & \textbf{OpenClaw}\enspace\seentag
      & \textbf{Hermes}\enspace\unseentag \\
    \midrule
    \rowcolor{copawgroupbg!45}
    Base model
      & 20.0
      & 24.5
      & 15.5 \\
    QwenPaw only
      & 29.1\enspace\gainbadge{9.1}
      & 27.3\enspace\gainbadge{2.8}
      & 22.7\enspace\gainbadge{7.2} \\
    OpenClaw only
      & 30.0\enspace\gainbadge{10.0}
      & 29.1\enspace\gainbadge{4.6}
      & 24.5\enspace\gainbadge{9.0} \\
    \addlinespace[3pt]
    \rowcolor{copawblue!5}
    \textbf{QwenPaw + OpenClaw}
      & 31.8\enspace\gainbadge{11.8}
      & 31.8\enspace\gainbadge{7.3}
      & 24.5\enspace\gainbadge{9.0} \\
    \bottomrule
  \end{tabularx}
  
\end{table}

Table~\ref{tab:cross-harness} shows that the benefits of our training data extend beyond the harness used for collection. Training on either source improves pass rates across all three evaluation harnesses, including Hermes, which is never used for data collection: QwenPaw-only and OpenClaw-only training yield absolute gains of \textbf{+7.2} and \textbf{+9.0} points on Hermes, respectively. Transfer is also evident between the two collection harnesses. In particular, OpenClaw-only training improves performance on QwenPaw by \textbf{+10.0 points}, exceeding the gain from QwenPaw-only training (+9.1 points) despite the mismatch between collection and evaluation harnesses. These findings suggest that the data teaches reusable capabilities for solving cowork tasks, with benefits that are not confined to harness-specific interaction patterns. Combining both sources achieves the highest pass rates on QwenPaw and OpenClaw (\textbf{31.8\%} on both), while retaining the full improvement on the unseen Hermes harness. Taken together, the results support the cross-harness generalizability of our data and its effectiveness in improving cowork task performance across different runtimes.

\subsubsection{Trajectory Selection}%
\label{sec:trajectory-filtering-results}
\leavevmode\par
\vspace{-0.5em}

\begin{table}[H]
  \centering
  \begin{threeparttable}
    \caption{Effect of model-conditioned trajectory filtering across model sizes. NLL and RSR each retain 70\% of the candidate trajectories.}%
    \label{tab:trajectory-filtering-v2}
    \small
    \setlength{\tabcolsep}{5pt}
    \renewcommand{\arraystretch}{1.16}
    \begin{tabularx}{\textwidth}{l*{3}{>{\centering\arraybackslash}X}}
      \toprule
      \rowcolor{copawgroupbg}
      \textbf{Selection}
        & \makecell[c]{\textbf{ClawEval}\\[-1pt]\textbf{Avg.}}
        & \makecell[c]{\textbf{PawBench}\\[-1pt]\textbf{Pass (\%)}}
        & \makecell[c]{\textbf{ClawMark}\\[-1pt]\textbf{Avg.}} \\
      \midrule
      \rowcolor{copawblue!5}
      \multicolumn{4}{l}{\textbf{Qwen3.5-9B}} \\
      \rowcolor{copawgroupbg!45}
      Base model & 0.6535 & 24.55 & 0.3276 \\
      No filter & \underline{0.7238} & 27.88 & 0.3267 \\
      NLL & \textbf{0.7295} & \textbf{29.39} & \textbf{0.4010} \\
      RSR & 0.7174 & \underline{28.48} & \underline{0.3991} \\
      \addlinespace[3pt]
      \rowcolor{copawblue!5}
      \multicolumn{4}{l}{\textbf{Qwen3.6-27B}} \\
      \rowcolor{copawgroupbg!45}
      Base model & 0.7311 & 39.09 & 0.6233 \\
      No filter & \underline{0.8065} & 39.09 & 0.6557 \\
      NLL & \textbf{0.8078} & \underline{43.03} & \textbf{0.6806} \\
      RSR & 0.7971 & \textbf{43.94} & \underline{0.6730} \\
      \addlinespace[3pt]
      \rowcolor{copawblue!5}
      \multicolumn{4}{l}{\textbf{Qwen3.6-35B-A3B}} \\
      \rowcolor{copawgroupbg!45}
      Base model & 0.7254 & 30.91 & 0.5344 \\
      No filter & 0.7667 & 39.39 & 0.5394 \\
      NLL & \textbf{0.7682} & \underline{40.91} & \underline{0.5678} \\
      RSR & \underline{0.7676} & \textbf{43.03} & \textbf{0.5739} \\
      \bottomrule
    \end{tabularx}
    \begin{tablenotes}[flushleft]
      \tiny
      \item Base model is the original checkpoint before post-training. No filter uses the same post-training recipe without NLL/RSR selection. NLL and RSR filter independently within ten equal-frequency log-length bins. Bold and underlined values denote the best and second-best result within each model size.
    \end{tablenotes}
  \end{threeparttable}
\end{table}

Retaining only 70\% of the evidence-valid pool, NLL selection improves every reported metric across all three model sizes. This result indicates that validity is necessary but does not make all trajectories equally valuable for learning. The preferred criterion also shifts with scale: NLL is strongest at 9B and 27B, whereas RSR performs best on most metrics at 35B-A3B. This reversal suggests that trajectory value is not intrinsic to the data alone, but depends on the capacity and predictive distribution of the target model. Since selection is performed within length bins, the effect cannot be readily attributed to length bias. A practical implication is that filtering criteria should be calibrated to the student model rather than applied uniformly across scales.

\subsection{On-Policy Distillation}%
\label{sec:opd-generalization-results}

\begin{figure}[H]
  \centering
  \includegraphics[width=\textwidth]{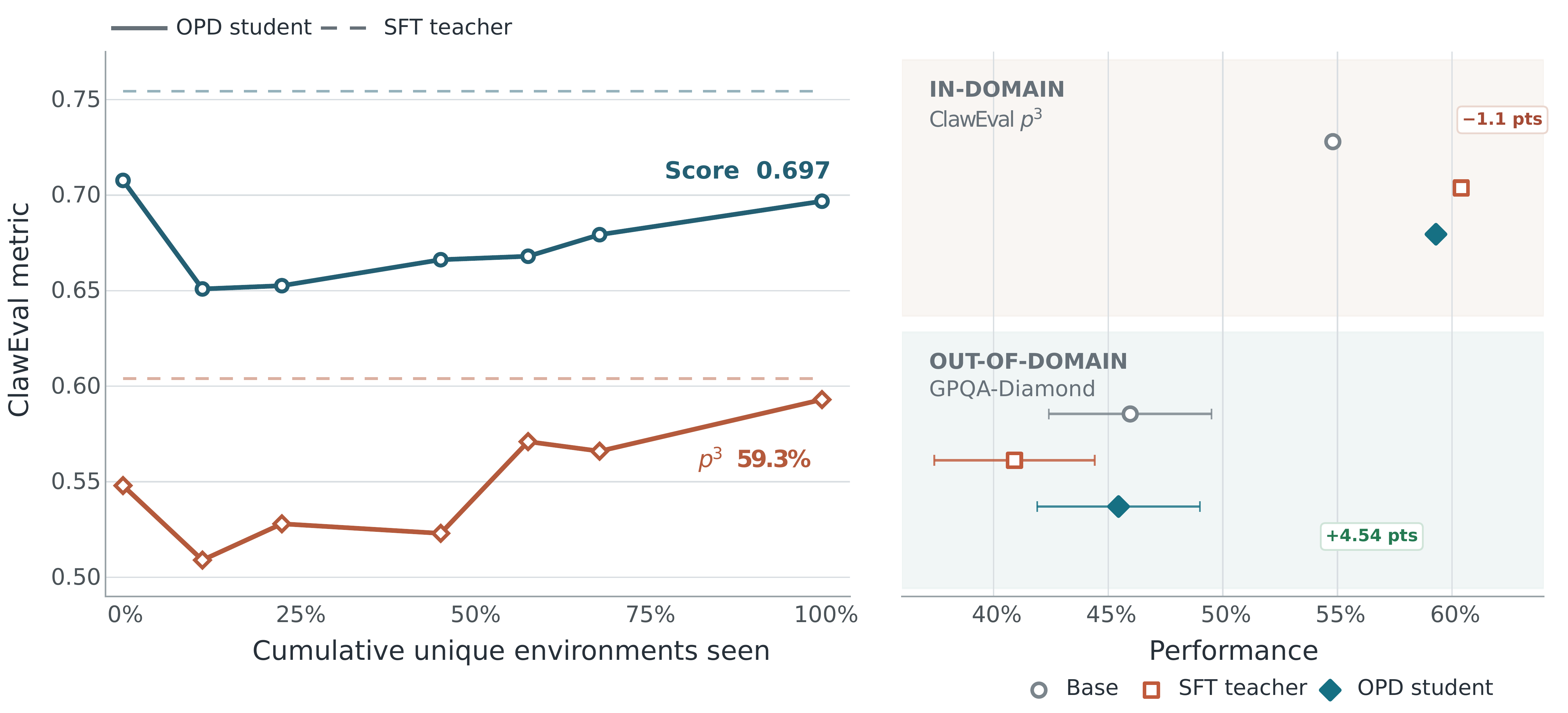}
  \caption{Agentic transfer and general-capability retention under OPD\@. Dashed lines mark SFT teacher performance; exposure is normalized to the final checkpoint.}%
  \label{fig:opd-environment-exposure}
\end{figure}

We finally examine whether agentic gains can be transferred while preserving broader model capabilities. On ClawEval, OPD improves $\text{pass}^3$ by \textbf{+4.5 points} over the base model, recovering approximately 80\% of the teacher's gain and leaving a gap of only 1.1 points. On GPQA-Diamond, the OPD student remains close to the base model while outperforming the SFT teacher by \textbf{+4.54 points}. This cross-benchmark pattern suggests that OPD transfers most of the specialized agentic capability while substantially limiting degradation in general performance.
\section{Related Work}%
\label{sec:related-work}

\paragraph{Executable agent environments.}
WebArena and OSWorld established reproducible, executable testbeds for web and computer-use agents \citep{webarena,osworld}. Recent work frames generalization in terms of scaling executable rules and state transitions, and automates environment discovery, task synthesis, validation, or solver-adaptive generation \citep{scalable_environments,agent_world,coevolve,agentevolver,clawenvkit,evoenv,verienv,scope,propel}. Complementary systems convert existing web, mobile, and desktop software into trainable environments \citep{gym_anything,webforge,weblica,shopgym,mobilegym}. Our pipeline applies these principles to personal work, jointly constructing heterogeneous workspaces, interaction interfaces, user policies, and outcome verifiers, with package-level validation before rollout.

\paragraph{Workspace and cowork evaluation.}
Workspace-Bench evaluates reasoning over large, heterogeneous file collections, while Agents' Last Exam and WildClawBench target economically grounded and long-horizon professional workflows \citep{workspace_bench,agents_last_exam,wildclawbench}. ClawMark further introduces multi-turn, evolving service state \citep{clawmark}. Claw-Eval and AJ-Bench emphasize trajectory, environment-state, and process evidence for reliable grading \citep{claw_eval,aj_bench}. These benchmarks characterize end-to-end agent systems; our work synthesizes the training environments and verified interaction data required to improve such systems.

\paragraph{Stateful user--agent interaction.}
ToolSandbox and $\tau$-bench combine tool execution with simulated multi-turn users and state-based evaluation \citep{toolsandbox,tau_bench}; $\tau^2$-Bench gives both participants tools over a shared environment \citep{tau2_bench}. VISTA extends user actions across UI and API interfaces \citep{vista}. Human-grounded studies find that general-purpose language models often simulate users with excessive cooperation and limited behavioral diversity, motivating data-grounded or explicitly optimized simulators \citep{sim2real_user_simulation,realusersim,turing_rl,odyssim}. We model task-conditioned interventions that can revise requirements, reject intermediate results, or mutate the shared workspace.

\paragraph{Agent post-training and harness effects.}
Interactive post-training combines online curricula, environment-grounded reinforcement learning, long-horizon trajectories, and on-policy distillation \citep{webrl,agent_r1,glm5,agents_a1,sdar,sdft}. Runtime research exposes context, memory, tools, scheduling, and access control as system-level services \citep{aios}; Harness-Bench and PawBench measure substantial variation across model--harness configurations \citep{harness_bench,pawbench}, while EvoTrainer co-evolves policies and training-side diagnostics \citep{evotrainer}. We treat the harness as an explicit rollout variable and derive SFT, RL, and distillation signals from validated environments and filtered trajectories.
\section{Limitations and Future Directions}%
\label{sec:limitations}

\paragraph{Task coverage.}
Our environments are synthetic approximations of personal work. Taxonomy-guided generation may omit tasks, cultures, languages, accessibility needs, and failure modes, while resource-grounded generation inherits the limits of available workspaces and interfaces. The current pool also emphasizes browser, file, data, and office workflows, leaving complex visual interfaces, specialized desktop software, and richer audio or video tasks underrepresented. Its scale should therefore not be interpreted as representative coverage. Future work should compare the synthesized distribution with real-world task distributions and extend multimodal state inspection and verification.

\paragraph{Verifier reliability.}
Verifier quality limits both data selection and reinforcement learning. Programmatic checks can be brittle, while language-model graders can be biased or overly permissive. Positive, negative, and mutation tests reduce but do not eliminate these risks, particularly for open-ended artifacts. Larger-scale human audits and calibration studies are needed to quantify false-acceptance and false-rejection rates.

\section{Conclusion}%
\label{sec:conclusion}

We introduced \frameworkname, a scalable framework for synthesizing executable Cowork environments. It combines taxonomy-guided and resource-grounded task generation with a build--execute--revise loop that constructs and validates the task, workspace, interfaces, user policy, and verifier as a complete environment. A stateful user simulator supports multi-turn collaboration through dialogue and shared-workspace edits. Using this framework, we synthesized 64.9K environments across 14 application scenarios and collected filtered trajectories for post-training.

Using these data, we post-trained 9B, 27B, and 35B-A3B Qwen models, all of which improve over their corresponding base checkpoints on average across six agentic benchmarks. Ours-27B ranks second overall and achieves the highest average scores on PawBench and ClawEval, outperforming evaluated models with more than 100B total parameters on both.

Scalable construction of executable environments can provide a data-driven path toward self-evolving agents. As tools, workspaces, and user requirements evolve, the environment pool can expand accordingly, supplying new tasks, interactions, and verification signals for subsequent training. This offers a path toward agents that improve continuously while adapting to changing work settings.

\clearpage
\bibliographystyle{style/iclr2026_conference}
\bibliography{references}

\clearpage
\appendix
\providecommand{\taxlabel}[1]{\nolinkurl{#1}}
\providecommand{\thead}[1]{\textbf{#1}}

\section{Task Taxonomy}%
\label{app:task-taxonomy}

The taxonomy-guided route uses a versioned taxonomy to control task generation and dataset coverage. The taxonomy separates the operations required by a task from its application domain and other task-level constraints. Labels describe the task to be constructed; they do not specify a solution trace or events observed during rollout.

\subsection{Taxonomy Overview}

The taxonomy contains six dimensions, summarized in \cref{tab:taxonomy-overview}. Capabilities and most task attributes may take multiple values, whereas each task is assigned one primary application path.

\begin{table}[H]
  \centering
  \caption{The six dimensions of the task taxonomy.}
  \label{tab:taxonomy-overview}
  \footnotesize
  \setlength{\tabcolsep}{4pt}
  \renewcommand{\arraystretch}{1.16}
  \begin{tabularx}{\linewidth}{@{}
    >{\raggedright\arraybackslash}p{0.22\linewidth}
    >{\raggedright\arraybackslash}p{0.16\linewidth}
    >{\raggedright\arraybackslash}X@{}}
    \toprule
    \rowcolor{copawgroupbg}
    \thead{Dimension} & \thead{Assignment} & \thead{What it controls} \\
    \midrule
    Capabilities & Multi-label & Operations expected to be required by the task. \\
    Application scenario & Single path & Primary domain and use case. \\
    Structure and difficulty & Structured attributes & Anticipated workflow, feedback, and task depth. \\
    User and deliverable & Structured attributes & Target user, output format, style, and length. \\
    Modality & Multi-label & Input and output channels. \\
    Safety and alignment & Multi-label & Relevant risks and expected response behavior. \\
    \bottomrule
  \end{tabularx}
\end{table}

\subsection{Capabilities}

Capability labels are independent of application scenarios. A task may therefore combine several operations within one domain without introducing a new domain-specific capability label.

\begingroup
\footnotesize
\setlength{\tabcolsep}{4pt}
\renewcommand{\arraystretch}{1.12}
\begin{longtable}{@{}
  >{\raggedright\arraybackslash}p{0.33\linewidth}
  >{\raggedright\arraybackslash}p{0.61\linewidth}@{}}
  \caption{Capability labels and their scope.}\label{tab:taxonomy-capabilities}\\
  \toprule
  \rowcolor{copawgroupbg}
  \thead{Capability} & \thead{Scope} \\
  \midrule
  \endfirsthead
  \multicolumn{2}{c}{\tablename~\thetable\ (continued)}\\
  \toprule
  \rowcolor{copawgroupbg}
  \thead{Capability} & \thead{Scope} \\
  \midrule
  \endhead
  \midrule
  \multicolumn{2}{r}{Continued on next page}\\
  \endfoot
  \bottomrule
  \endlastfoot
  \taxlabel{Logic_Reasoning} & Causal, deductive, abductive, and constraint-based reasoning. \\
  \taxlabel{Math_Scientific} & Arithmetic, statistical, and scientific calculation. \\
  \taxlabel{Code_Manipulation} & Code generation, debugging, review, refactoring, and testing. \\
  \taxlabel{Tool_Calling} & Tool selection, argument construction, and chained tool use. \\
  \taxlabel{Knowledge_Retrieval} & Evidence extraction, cross-document comparison, and retrieval verification. \\
  \taxlabel{Web_Browsing} & Web search, page retrieval, extraction, and interaction. \\
  \taxlabel{Data_Analysis} & Data cleaning, analysis, visualization, and reporting. \\
  \taxlabel{Content_Generation} & Composition and synthesis of written content. \\
  \taxlabel{Multimodal_Perception} & Interpretation of image, video, and audio inputs. \\
  \taxlabel{Multimedia_Processing} & Editing or transformation of image, video, and audio assets. \\
  \taxlabel{Calendar_Scheduling} & Calendar operations and temporal coordination. \\
  \taxlabel{Email_Communication} & Email retrieval, classification, drafting, and sending. \\
  \taxlabel{File_Management} & File operations, conversion, archiving, and directory management. \\
  \taxlabel{Security_Awareness} & Recognition and mitigation of injection, leakage, and social-engineering risks. \\
  \taxlabel{Agent_Self_Config} & Configuration of agents, skills, personas, and scheduled tasks. \\
  \taxlabel{Translation} & Translation and cross-lingual information alignment. \\
  \taxlabel{Spatial_Visual_Reasoning} & Spatial inference, visual puzzles, and geometric reconstruction. \\
\end{longtable}
\endgroup

\subsection{Application Scenarios}

Each task is assigned one primary scenario path. The hierarchy has three levels: a broad domain, a subdomain, and a specific use case. \Cref{tab:taxonomy-scenarios} summarizes the top two levels and gives one representative leaf for each domain.

\begingroup
\scriptsize
\setlength{\tabcolsep}{4pt}
\renewcommand{\arraystretch}{1.16}
\begin{longtable}{@{}
  >{\raggedright\arraybackslash}p{0.27\linewidth}
  >{\raggedright\arraybackslash}p{0.41\linewidth}
  >{\raggedright\arraybackslash}p{0.26\linewidth}@{}}
  \caption{Top-level application scenarios, subdomains, and representative leaves.}\label{tab:taxonomy-scenarios}\\
  \toprule
  \rowcolor{copawgroupbg}
  \thead{Scenario} & \thead{Subdomains} & \thead{Example leaf} \\
  \midrule
  \endfirsthead
  \multicolumn{3}{c}{\tablename~\thetable\ (continued)}\\
  \toprule
  \rowcolor{copawgroupbg}
  \thead{Scenario} & \thead{Subdomains} & \thead{Example leaf} \\
  \midrule
  \endhead
  \midrule
  \multicolumn{3}{r}{Continued on next page}\\
  \endfoot
  \bottomrule
  \endlastfoot
  \taxlabel{Office_Productivity} & Email, Calendar, Document, Meeting, Task Management, Expense & \taxlabel{Inbox_Triage} \\
  \taxlabel{Information_Retrieval} & Academic, Market, News, Social Platform, Knowledge Base & \taxlabel{Literature_Review} \\
  \taxlabel{Finance_Investment} & Stock, Financial Report, Forex, Investment Research & \taxlabel{Earnings_Analysis} \\
  \taxlabel{Software_Engineering} & Code, Frontend, DevOps, GitHub, Database & \taxlabel{Debugging} \\
  \taxlabel{Data_Science} & Business Intelligence, Financial Analysis, Text Analytics & \taxlabel{KPI_Dashboard} \\
  \taxlabel{Content_Media} & Writing, Social Media, Design, Video and Audio & \taxlabel{Academic_Writing} \\
  \taxlabel{Safety_Compliance} & Prompt Injection, Data Protection, Harmful Content, User Wellbeing & \taxlabel{Content_Injection} \\
  \taxlabel{Automation_Platform} & Agent, Skill, Scheduling & \taxlabel{Skill_Installation} \\
  \taxlabel{Travel_Lifestyle} & Travel, Daily Life & \taxlabel{Itinerary_Planning} \\
  \taxlabel{Education_Knowledge} & Learning, Knowledge QA, Exam Preparation & \taxlabel{Concept_Explanation} \\
  \taxlabel{Legal} & Contract, Litigation & \taxlabel{Clause_Extraction} \\
  \taxlabel{Healthcare} & Medical Policy, Pet Care & \taxlabel{Policy_QA} \\
  \taxlabel{Manufacturing_Engineering} & Quality Control, Aerospace, Smart Agriculture & \taxlabel{Defect_Detection} \\
  \taxlabel{Multimodal_Generation} & Webpage, score, map, video, and document transformations & \taxlabel{Video_To_Webpage} \\
\end{longtable}
\endgroup

\section{Data Sources}
\label{app:data-sources}

Our environment file pool draws on diverse public and versioned sources;
representative examples are listed below.

\begin{table}[H]
  \centering
  \caption{Representative sources used to construct the environment file pool.}
  \label{tab:data-sources}
  \scriptsize
  \setlength{\tabcolsep}{4pt}
  \renewcommand{\arraystretch}{1.12}
  \begin{tabularx}{\linewidth}{@{}
    >{\raggedright\arraybackslash}p{0.22\linewidth}
    >{\raggedright\arraybackslash}X@{}}
    \toprule
    \rowcolor{copawgroupbg}
    \thead{Source type} & \thead{Representative sources} \\
    \midrule
    Hugging Face datasets &
      \nolinkurl{superdoc-dev/docx-corpus}; WikiArt; EuroSAT; MathVista;
      Food-101; \nolinkurl{common-pile/arxiv_papers}; Zenodo10K; Enron Corpus;
      LibriSpeech; ESC-50; UrbanSound8K; and
      MINDS-14 \\
    Regulatory and disclosure portals &
      SEC EDGAR, CNINFO, and CSE filings \\
    Government open data &
      \nolinkurl{data.gov.hk}; national portals for the United Kingdom,
      Australia, Ireland, and Italy; and the Toronto CKAN catalog \\
    Fixed GitHub revisions &
      Zeek, Lucide, Simple Icons, MutopiaProject, music21, and
      OpenSheetMusicDisplay \\
    Vertical-domain websites &
      \nolinkurl{ypppt.com} and domain-specific public-sector sites such
      as \nolinkurl{bjfsh.gov.cn} \\
    \bottomrule
\end{tabularx}
\end{table}

\section{Prompts for Environment Construction and Quality Control}
\label{app:agent_prompts}

We provide abridged versions of the prompts used by the three major
agents in our environment-construction pipeline. The prompts preserve
the main role definitions, directory structure, runtime interfaces,
workspace construction rules, user-simulator design, verifier
construction, and quality-control procedures, while omitting repeated
engineering instructions and deployment-specific special cases.

\newenvironment{compactprompt}[1]
{\VerbatimEnvironment
 \begin{tcolorbox}[
   enhanced jigsaw,
   breakable,
   lines before break=6,
   width=\linewidth,
   colback=copawpromptbg,
   colframe=copawpromptframe,
   colbacktitle=copawprompttitlebg,
   coltitle=copawblue!75!black,
   title={#1 Prompt},
   title after break={#1 Prompt\hfill\normalfont\scriptsize\itshape continued},
   fonttitle=\footnotesize\bfseries,
   boxrule=0.5pt,
   leftrule=2pt,
   arc=1.2mm,
   outer arc=1.2mm,
   boxsep=0pt,
   left=1.5mm,
   right=1.5mm,
   top=0.8mm,
   bottom=0.8mm,
   pad at break*=0.8mm,
   before skip=0.5\baselineskip,
   after skip=0.5\baselineskip]
 \begin{Verbatim}[fontsize=\scriptsize,
                  breaklines=true,
                  breakanywhere=true,
                  breaksymbolleft={},
                  breaksymbolright={},
                  baselinestretch=0.82,
                  xleftmargin=0pt,
                  xrightmargin=0pt]}
{\end{Verbatim}
 \end{tcolorbox}}

\subsection{Builder}
\label{app:builder-prompt}

The Builder instantiates an executable sample from the task
specification. It constructs the visible workspace, configures the
runtime tools and skills, prepares the private user-simulator context,
and designs multi-turn or collaborative workspace updates. It also
executes the task once to obtain a grounded reference answer, but does
not construct the verifier.

\begin{compactprompt}{Builder}
You are a sample Builder. Construct an executable agentic sample in the current task root.

Construct the visible workspace and task inputs; configure runtime tools, skills, metadata, and user messages; prepare private user-simulator and collaborative-update state; execute the task to derive a grounded reference answer; then restore the intended initial state. Do not implement verifier logic or modify tests/.

Use only configured runtime interfaces. The runtime allowlist includes execute_shell_command; file read, write, edit, listing, glob, and grep operations; browser_use and download_url; image/video inspection; time, token, collaboration, memory, and Tavily retrieval tools. Builder-only generate_image, generate_video, generate_music, and generate_speech tools must not appear in the execution agent's requirements.

Construct real, usable visible inputs under:
  environment/data/local_files/
    workspace/
    downloads/
    notes/
    references/

Expose only runtime paths such as workspace/..., downloads/..., notes/..., and references/...; never expose construction paths. Maintain state_manifest.json for organization and provenance without answers, expected classifications, exact values, verifier details, or hidden evidence. Do not pre-create the deliverable. Multimodal and GUI tasks must provide accessible assets or operable interfaces and require meaningful observation or interaction rather than direct source-file shortcuts.

For multi-turn or collaborative tasks, construct private context under:
  .user/
    persona.md
    profile.json
    domain_knowledge.md
    world_model.md
    private_rules.md
    memory/..., history/..., state/...

Follow the supplied schemas. Private context must model a coherent user's knowledge, terminology, tools, constraints, and plausible blind spots without encoding the task or solution. Later turns must introduce new information, changed inputs or requirements, or decisions based on prior progress rather than splitting deterministic cleanup steps.

For collaborative tasks, create .patch/instructions.md and turn_NN_hint.md files using supported create, overwrite, append, delete, or edit actions. Patches must use runtime-visible paths, satisfy action preconditions, and materially change subsequent work.

Update task.yaml with the required metadata, labels, tools, skills, environment paths, setup scripts, messages, and evaluation fields while preserving pipeline identifiers. When applicable, configure successful/not_successful, edit, undo, sentiment, behavior, or artifact-download signals only when they affect active interaction state.

Execute the task on the real inputs and derive values, records, summaries, artifacts, and state changes rather than guessing. Store a non-placeholder reference answer and, when applicable, answer_facts for commands, files, values, and GUI state. For multi-turn tasks, organize the answer as Round 1, Round 2, ....

Difficulty should arise from meaningful dependencies, cross-format reasoning, tool use, revisions, or conflicting evidence. Before finishing, inspect all visible files; remove logs, caches, debug artifacts, temporary screenshots, and reference-run outputs; restore the initial workspace; and confirm that no hidden answer or verifier information is exposed.

...
\end{compactprompt}

\subsection{Verifier Constructor}
\label{app:verifier-constructor-prompt}

The Verifier Constructor receives the constructed task and independently
builds the evaluation logic. It derives expected values from the task
inputs, combines deterministic checks with trajectory-level rubric
criteria, and self-validates the verifier on both correct and controlled
failure states.

\begin{compactprompt}{Verifier Constructor}
You are the Verifier Constructor for an agentic task.

Your responsibility is verification only. Do not redesign the task, workspace, user context, or inputs; report inconsistencies to the Builder.

Construct:
  tests/
    checklist.jsonl
    verification.py
    verify.py
    test.sh
    test_outputs.py

Read task.yaml, the real files under environment/data/local_files/, evaluation.inputs.answer, answer_facts, and relevant multi-turn or workspace-update state. Treat reference claims as hypotheses: first reproduce the key expected results from the inputs.

For each user requirement, build:
  requirement
    -> expected outcome
    -> observable evidence
    -> checker or rubric item
    -> weight

Prefer deterministic, format-aware checks for file contents, JSON/CSV/XLSX fields, numeric results, schemas, sets, generated artifacts, cross-file consistency, and observable state changes. Use numeric tolerances; normalize dates, currencies, and percentages; and use case-insensitive or regex matching where wording is flexible. Avoid checks tied to incidental phrasing, CSS classes, pixel positions, or one implementation when equivalent outputs are valid.

Cover output validity, structural correctness, semantic correctness, and red-line constraints, with semantic and safety requirements weighted above file existence. Inspect trajectory evidence only when completion requires a particular process, such as viewing required media, interacting through browser_use, or re-reading an updated source before recomputation.

Create tests/checklist.jsonl with entries:
{
  "id": "...",
  "item": "...",
  "weight": ...,
  "status": "pending",
  "evidence": ""
}

Checklist items must be task-relevant trajectory assertions using real runtime tool names and observable evidence, not static package checks. Typical categories cover required tools, observed inputs, required procedure, final-answer facts, and shortcut avoidance.

Emit structured results containing passed, score, max_score, score_points, summary, and errors. Missing outputs, malformed artifacts, or absent trajectory evidence must yield a normal low score rather than an exception.

Reject untouched states, empty or copied outputs, unsupported claims, pre-existing artifacts presented as completed work, and task-specific shortcuts. For in-place editing, verify that the requested change occurred.

Self-validate using a correct completion, the untouched state, missing or malformed outputs, wrong key values, cross-file inconsistencies, missing trajectory evidence, and relevant shortcuts. Do not weaken core checks to make the reference pass; locate defects in the task, data, reference, or verifier and report them to the responsible constructor.

...
\end{compactprompt}

\subsection{Issue Detector}
\label{app:issue-detector-prompt}

The Issue Detector performs the final package-level quality review. It
actively inspects the environment, reproduces the task, checks multi-turn
and workspace-update behavior, and audits whether the verifier
distinguishes valid completions from incomplete or shortcut states. It
does not repair the package directly; instead, it returns evidence-based
defects to the responsible constructor.

\begin{compactprompt}{Issue Detector}
You are an active quality Issue Detector.
Evaluate the sample before release. Use tools to inspect real files, reproduce the task, and test the verifier; do not judge from prompt text alone and do not repair the package directly.

1. Inspect the package
Read the complete task.yaml; environment/data/local_files/; optional .user/, .patch/, and solution/ artifacts; tests/checklist.jsonl; and the verifier entry points and implementation. Inspect actual contents rather than filenames.

Evaluate the execution agent under its real visibility boundary. Hidden user state, future patches, solutions, tests, and construction metadata are shortcuts only if leaked into runtime-visible state. Check visible inputs for real usable resources, placeholders, answer or verifier leakage, private-state exposure, and construction residue.

2. Inspect user context and patches
For multi-turn tasks, verify that persona, profile, knowledge, memory, history, and state describe a coherent user; schemas are valid; domain knowledge is task-relevant rather than a solution; and latent goals do not expose the task.

For collaborative tasks, validate patch YAML, allowed actions, runtime-visible paths, create/edit/overwrite preconditions, and cross-turn consistency. Workspace changes must materially affect later work; empty patches or explanatory notes alone are insufficient.

3. Reproduce the task
Actually perform the task using the supplied inputs and allowed tools. Parse structured data, inspect documents and required media, and interact with GUI state when applicable. Record tools used, files observed, important values, intermediate calculations, generated artifacts, and final conclusions.

Compare the reproduction with evaluation.inputs.answer and answer_facts, and report exact mismatches. For each later turn, verify that it depends on new information, changed inputs or requirements, user preferences, workspace updates, or actual prior results; reject turns that merely split deterministic work or request confirmation.

4. Review the checklist
Each applicable item must use real runtime tool names, refer to observable evidence, match the task, be satisfiable by a valid execution, and avoid imposing one arbitrary implementation. Identify missing anti-shortcut coverage, but do not modify the checklist.

5. Audit the verifier

Inspect the mapping:
  requirement
    -> success condition
    -> evidence
    -> checker
    -> score weight

Audit layered scoring, deterministic priority, task specificity, rigor, and anti-cheating. Semantic correctness and red-line constraints must outweigh superficial file-existence checks.

Run the verifier, when possible, on the untouched state, a correct reference completion, and controlled failures such as missing or malformed outputs, wrong key values, cross-file inconsistencies, absent trajectory evidence, and task-specific shortcuts. It must reject incomplete, incorrect, untouched, and shortcut states without crashing; passing the initial state is a serious defect.

6. Score and route
Score realism, separation, solvability, interaction_potential, consistency, and patch_quality when applicable, then compute the applicable average. Critical executability, leakage, and verifier failures override a high average.

Use repair targets task_spec, environment, solution, verifier, trajectory, or packaging. Task, environment, user-context, and reference-answer defects normally return to the Builder; verifier defects return to the Verifier Constructor. Do not recommend weakening core checks merely to force a pass.

7. Final output
Return exactly one JSON object:
{"realism": 0, "separation": 0, "solvability": 0,
 "interaction_potential": 0, "consistency": 0, "patch_quality": 0,
 "average": 0.0, "passed": false,
 "fail_reasons": ["..."], "repair_suggestions": ["..."],
 "repair_targets": ["..."]}
Failure reasons must cite concrete evidence. Set passed=true only when the quality threshold and all hard executability and verification checks pass.

\end{compactprompt}

These prompts summarize method-level instructions. Runtime-specific interface,
sandbox, and execution fragments are omitted for space without altering the
roles or pipeline.

\end{document}